\documentclass[10pt, reqno]{article}
\usepackage{graphicx}
\usepackage{indentfirst,csquotes}
\usepackage[margin=1in]{geometry}

\usepackage{amssymb,amsthm,amsmath}
\usepackage{xcolor,paralist,hyperref,fancyhdr,etoolbox}
\usepackage{lipsum}
\usepackage{forest}
\usepackage[algoruled, lined, linesnumbered, noend]{algorithm2e}
\usepackage[numbers]{natbib}
\usepackage{booktabs}
\usepackage{multirow}
\usepackage{authblk}
\usepackage{pdflscape}

\newtheorem{rem}{Remark}

\hypersetup{
  colorlinks=true, 
  linkcolor=blue, 
  filecolor=black, 
  urlcolor=black, 
  citecolor=blue
}

\newcommand{\E}{\mbox{E}}

\title{From Point to probabilistic gradient boosting for claim frequency and severity prediction}
\author[1]{Dominik Chevalier}
\author[1]{Marie-Pier Côté}
\affil[1]{École d'actuariat, Université Laval, 2425, rue de l'Agriculture, Québec, QC, Canada, G1V~0A6.}
\affil[ ]{Corresponding author: Dominik Chevalier (\href{mailto:dominik.chevalier.1@ulaval.ca}{dominik.chevalier.1@ulaval.ca})}
\date{\today}

\begin{document}
\maketitle

\begin{abstract}
Gradient boosting for decision tree algorithms are increasingly used in actuarial applications as they show superior predictive performance over traditional generalised linear models. Many enhancements to the first gradient boosting machine algorithm exist. We present in a unified notation, and contrast, all the existing point and probabilistic gradient boosting for decision tree algorithms: GBM, XGBoost, DART, LightGBM, CatBoost, EGBM, PGBM, XGBoostLSS, cyclic GBM, and NGBoost. In this comprehensive numerical study, we compare their performance on five publicly available datasets for claim frequency and severity, of various sizes and comprising different numbers of (high cardinality) categorical variables. We explain how varying exposure-to-risk can be handled with boosting in frequency models. We compare the algorithms on the basis of computational efficiency, predictive performance, and model adequacy. LightGBM and XGBoostLSS win in terms of computational efficiency. CatBoost sometimes improves predictive performance, especially in the presence of high cardinality categorical variables, common in actuarial science. The fully interpretable EGBM achieves competitive predictive performance compared to the black box algorithms considered. We find that there is no trade-off between model adequacy and predictive accuracy: both are achievable simultaneously.$^{2, 3}$
\end{abstract} 
\maketitle
\section{Introduction}\label{sec1}

While generalised linear models (GLMs) have been the cornerstone of general insurance ratemaking for many years \cite{DeJong/etal:2008:GLMInsurance,frees:2015}, recent advancements in machine learning provide actuaries with a wide range of competing predictive algorithms. In particular, gradient boosting machine (GBM) can improve predictive performance over GLMs for risk segmentation \cite{Henckaerts/etal:2021}. Gradient boosting proliferates in actuarial applications, such as auto insurance \citep{Clemente/etal:2023, Gao/etal:2023}, hierarchical non-life reserving \citep{Crevecoeur/etal:2022}, and health insurance \citep{Hartman/etal:2020, Hancock/etal:2020}.

GBM, proposed by Friedman \cite{Friedman:2001}, is a predictive algorithm that combines weak learners, simple models that are usually decision trees, to gradually improve predictions by leveraging gradient descent, an optimisation technique. The enhanced precision of GBM comes with the cost of an opaque computationally intensive algorithm. Many refinements of gradient boosting for decision trees (GBDT) tackle computational efficiency, such as XGBoost \citep{Chen/etal.:2016:XGBoost} and LightGBM \citep{Ke/etal.:2017:LightGBM}, the treatment of categorical variables, such as CatBoost \citep{Prokhorenkova/Etal.:2018:Catboost}, and model interpretability, such as explainable GBM (EGBM) \citep{lou:2012:intelligibleCAR, lou:2013:GA2M}. 

The above GBDT algorithms perform point prediction, which, in most cases, leads to an estimator of the expected response variable conditional on covariates. They do not describe the general behaviour of the response variable and its distribution, crucial elements for risk management purposes in actuarial modelling. As an alternative, in their ``recent challenges in actuarial science'', Embrechts and W\"uthrich~\cite{embrechts:2022} recommend transitioning towards probabilistic prediction for a better uncertainty assessment. In~\cite{Meng/etal:2022}, this idea is used for the construction of a dual-parameter tree boosting method to emulate the GLM framework with zero-inflated data, an approach generalized in cyclic GBM (cyc-GBM) \cite{Delong/etal:2023}. Probabilistic gradient boosting techniques were applied in the context of zero-inflated insurance claim frequency  \cite{So/Valdez:2024,Power/etal:2023}, and along with copula models for multivariate claim severities \cite{Power/etal:2023}.

From the machine learning angle, Breiman \cite{breiman:2001} opposes the ``Data modelling culture'', in which we try to understand the probabilistic distribution of the data, to the ``Algorithmic modelling culture'', in which we aim to estimate accurate predictor functions regardless of distributional assumptions. The line between the two cultures is blurry with the rise of probabilistic forecasting models: motivated by reconciling these cultures, M\"arz \cite{Marz:2019:xgboostlss} proposes XGBoostLSS, that sequentially runs XGBoost algorithms for multi-parameter predictions. This is the first of many probabilistic GBDT algorithms, whose objective is to predict distributional parameters given covariates. In the same vein, Delong et al. \cite{Delong/etal:2023} take advantage of the cyclic coordinate descent optimisation algorithm in cyc-GBM. Because the gradient is not invariant to reparametrisation, Duan et al. \cite{Duan/etal.:2020:NGBoost} replace it by a rescaled version in their natural gradient boosting (NGBoost) algorithm. Predicting the expectation and variance is the focus in the probabilistic GBM algorithm PGBM \cite{sprangers2021probabilistic}, built in the XGBoost framework.

In this comprehensive review, we present in a unified notation the point and probabilistic GBDT algorithms, and compare their performance in a numerical study on a variety of insurance datasets. We explain the passage from a point prediction to a predicted distribution per observation. We examine five point prediction GBDT approaches (GBM, XGBoost, LightGBM, CatBoost, and EGBM) and four probabilistic algorithms (XGBoostLSS, cyc-GBM, NGBoost and PGBM). We show how varying exposure-to-risk can be handled in GBDT frequency models. We also perform a numerical study on claim frequency and severity data with five publicly available datasets resulting in a novel impartial comparative analysis. The angles of comparison are  computational efficiency, predictive performance, portfolio balance and calibration, model adequacy, and tariff structures. We provide results for GLMs and generalised additive models for location, shape, and scale (GAMLSS) to contrast GBDT with traditional approaches.  We find that LightGBM stands out as the most computationally efficient with little or no loss in predictive performance. Amongst the probabilistic algorithms, XGBoostLSS is the most computationally efficient while providing adequate fit in terms of the coverage of confidence intervals. Interestingly, the probabilistic GBDT algorithms can enhance model adequacy without hindering predictive performance. Although all algorithms achieve similar predictive performance, CatBoost can lead to a small improvement when the dataset contains high cardinality categorical variables. In terms of portfolio balance  and calibration, we find that LightGBM and cyc-GBM, respectively, can perform worse than other GBDT models.

Probabilistic GBDT refinements are developed concurrently and introduced in the literature with different notations, making it difficult for users to grasp their similarities and particularities. Our contribution is two-fold. First, our comprehensive review presents the probabilistic GBDT algorithms in a coherent and consistent manner. Second, our numerical study on claim frequency and severity data is impartial and sheds light on computational efficiency, predictive performance, balance, calibration, segmentation, and model adequacy, the latter being rarely studied with GBDT. We use five datasets and ten GBDT algorithms, some of which have never been compared. 

The remainder of this paper is structured as follows. To establish the basis of the two types of GBDT algorithms, we distinguish point and probabilistic predictions in Section~\ref{sec:frompointtoprob}. Then, we describe the GBDT algorithms in Section~\ref{sec:survey}. The experimental setup and model evaluation framework are presented in Section~\ref{sec:methodology}. Section~\ref{sec:results} contains the numerical comparative study, and a discussion follows in Section~\ref{sec:discussion}. The tuning procedure is detailed in Appendix~\ref{app:tuning} and additional results are available in Appendix~\ref{app:addres}.

\section{From point to probabilistic prediction}
\label{sec:frompointtoprob}

Let $Y$ be a response variable, i.e., the target, and $\mathbf{x}$ be the associated $\eta$-dimensional vector of covariates. Let $\mathcal{D}=\{(\mathbf{x}_i, y_i) : i=1,\ldots,n\}$ be a dataset of $n$ independent copies of $(\mathbf{x},Y)$. GBDT algorithms solve a supervised learning task and rely on a pre-specified loss function. For observation $i\in\{1,\ldots,n\}$, the loss function $\mathcal{L}(y_i, b_i)$ takes two arguments: the observed target $y_i$ and a candidate prediction $b_i$ that may depend on $\mathbf{x}_i$. The GBDT training algorithms iteratively update the candidate predictions~$b_1, \ldots, b_n$, in an attempt to minimise the total loss for $\mathcal{D}$, defined as $\sum_{i=1}^n \mathcal{L}(y_i, b_i)$. The possibility of choosing any loss function makes gradient boosting versatile.
 
The squared error loss $\mathcal{L}(y_i, b_i) = (y_i - b_i)^2$ is commonly used to tackle regression tasks.
Optimising the total squared error loss amounts to an ordinary least squares (OLS) procedure or a maximum likelihood estimation (MLE) of the location parameter of an elliptical distribution (e.g., a Gaussian distribution). However, because symmetric distributions are rarely appropriate for claim severity and frequency modelling, other loss functions like the gamma or Poisson deviances should be considered \citep{wuthrich:2023}. For observation $i$, the gamma deviance is given by 
$$
\mathcal{L}(y_i, b_i) \propto\left(\frac{y_i - b_i}{b_i} - \ln \frac{y_i}{b_i}\right),
$$
and the Poisson deviance is
$$
\mathcal{L}(y_i, b_i) \propto \left\{y_i \ln \left(\frac{y_i}{b_i}\right) - (y_i - b_i ) \right\},
$$
where we impose that $y_i \ln y_i = 0$ if $y_i=0$ \citep[see, e.g.,][]{denuit/etal:2019}. The family for the conditional distribution of $Y$ is implied by the choice of loss function. Like for the squared error loss, which can be viewed as the Gaussian deviance, the minimum of these deviances is attained at $b_i = \E[Y_i|\mathbf{x}_i]$. Therefore, the learned prediction function when the objective is the total deviance can be interpreted as an estimator $\hat{\mu}(\mathbf{x})$ of $\E[Y|\mathbf{x}]$. It would be possible to specify a loss function to perform quantile regression as seen, e.g., in~\cite{fenske/etal:2011,velthoen:2023}, but we focus on point predictions for the conditional expectation.

In the OLS framework, after the estimation of the regression coefficients, we get point predictions for $\hat{\mu}(\mathbf{x})$. Given these $\hat{\mu}(\mathbf{x}_1), \ldots, \hat{\mu}(\mathbf{x}_n)$, the (constant) variance parameter can be estimated by maximising the log-likelihood under the assumption of homoscedasticity. Similarly, given predicted means $\hat{\mu}(\mathbf{x}_1), \ldots, \hat{\mu}(\mathbf{x}_n)$ from a GBDT algorithm, other (constant) distributional parameters can be estimated by maximising the log-likelihood under a homogeneity assumption. This last assumption means that scale and/or shape parameters do not vary with $\mathbf{x}$. This way, we obtain a predicted distribution per observation, leveraging point prediction to build probabilistic models.

To relax the homogeneity and linearity assumptions in GLMs, Rigby and Stasinopoulos \cite{rigby/stasinopoulos:2005} propose GAMLSS. They allow all the parameters of the conditional distribution of $Y$ given~$\mathbf{x}$ to be modelled by smooth functions of the covariates. GAMLSS equip us for ``more flexible modelling of more suitable distributions'' \cite{turcotte/Boucher:2024} in the actuarial context, so it is natural to combine the flexibility of GAMLSS with the predictive potential of GBDT. To this end, probabilistic GBDT algorithms can predict all the distributional parameters as functions of $\mathbf{x}$, thereby granting them with the advantages of GAMLSS noted above. In addition, most distributions may be parametrised by a mean (often the location) parameter, whose estimate corresponds to a point prediction. Thus, probabilistic GBDT algorithms yield point predictors as a by-product of the estimated distribution.

\section{Survey of gradient boosting for decision trees}
\label{sec:survey}

Gradient boosting stems from Schapire's ``strength of weak learnability'' \cite{Schapire:1990}, a theory in which the iterative summation of simple models, called weak learners, can form strong predictive models. Friedman \cite{Friedman:2001} proposes the classification and regression trees (CART) of Breiman et al. \cite{Breiamn:1984} as weak learners in a gradient descent procedure. In such a GBDT algorithm, we aggregate a predetermined number $M$ of simple decision trees to obtain the flexible GBDT predictor function
$$
f_{\mathrm{GBDT}}^{M}(\mathbf{x}) = f_{\mathrm{GBDT}}^{M - 1}(\mathbf{x}) + h_M(\mathbf{x})= \sum_{m =  1}^{M} h_m(\mathbf{x}),
$$
where $h_m(\mathbf{x})$ is the prediction function of the decision tree in the $m$th iteration. Some enhancements follow Friedman's method \cite{Friedman:2001}: they tackle its lack of transparency, the computational weaknesses of the CART weak learners, and probabilistic prediction. Fig.~\ref{fig:implementations} is a comprehensive concept map of the implementations of GBDT. Note that we only consider gradient boosting, and not other types such as Delta boosting \cite{lee:2018}.

\smallskip
\begin{rem}
 When the domain of the target parameter, say $\mu$, is not $\mathbb{R}$, we introduce a link function~$\varphi$ to properly restrict the prediction to the desired set. We express the loss function $\mathcal{L}(y,b)$ such that $b = \varphi(\mu) \in \mathbb{R}$ under the chosen link function. The output prediction function of the GBDT is then 
\begin{align*}
\hat{\mu}(\mathbf{x}) = f_{\mathrm{GBDT}}(\mathbf{x}) = \varphi^{-1}\{f_{\mathrm{GBDT}}^{M}(\mathbf{x})\}.
\end{align*}
It is common to use the log link $\varphi(x)=~\log(x)$ for a positive target parameter.
\end{rem}
\smallskip

In this section, we review GBDT algorithms. In Section~\ref{sec:gbm}, we describe point prediction GBDT algorithms, including computational enhancements and the issue of model interpretability. Probabilistic GBDT methods are presented in Section~\ref{sec:probalgos}: XGBoostLSS, cyc-GBM, and NGBoost. We discuss hybrid strategies in Section~\ref{sec:hybrid}.

\begin{figure}[!ht]  
    \centering
 \begin{forest}
        box/.style={rounded corners, draw},
        for tree={s sep=4mm}
        [\begin{tabular}{@{}c@{}}GBM \cite{Friedman:2001, Friedman:2002:StochasticGradient} \end{tabular}, box, 
            [\begin{tabular}{@{}c@{}}EGBM \cite{lou:2012:intelligibleCAR,lou:2013:GA2M} \end{tabular}, box, anchor = south]
            [\begin{tabular}{@{}c@{}}DART  \cite{Rashimi:2015} \end{tabular}, box]
            [\begin{tabular}{@{}c@{}} XGBoost \cite{Chen/etal.:2016:XGBoost} \end{tabular}, box
                    [\begin{tabular}{@{}c@{}}LightGBM \cite{Ke/etal.:2017:LightGBM} \end{tabular}, box
                        [\begin{tabular}{@{}c@{}}CatBoost \cite{Prokhorenkova/Etal.:2018:Catboost} \end{tabular}, box]]
                        [\begin{tabular}{@{}c@{}}XGBoostLSS \cite{Marz:2019:xgboostlss} \\ \end{tabular}, box, fill = black!20,
                        [\begin{tabular}{@{}c@{}}XGBoostLSS DART \cite{Meng/etal:2022} \\ \end{tabular}, box, fill = black!20]]
                        [\begin{tabular}{@{}c@{}} PGBM\cite{sprangers2021probabilistic} \end{tabular}, box, fill = black!20]
                        ]
            [\begin{tabular}{@{}c@{}} NGBoost\cite{Duan/etal.:2020:NGBoost} \end{tabular}, box, fill = black!20]
           [\begin{tabular}{@{}c@{}} cyc-GBM\cite{Delong/etal:2023} \end{tabular}, box, fill = black!20, anchor = south]
        ]
    \end{forest}
    \caption{Comprehensive concept map of point (white) and probabilistic (grey) GBDT algorithms.}
    \label{fig:implementations}
\end{figure}

\subsection{Gradient boosting machines and their enhancements for point prediction}
\label{sec:gbm}

Algorithm~\ref{alg:gbm} presents stochastic gradient boosting machine (GBM), the root of the concept map in Fig.~\ref{fig:implementations}, introduced in \cite{Friedman:2001, Friedman:2002:StochasticGradient}. The way we present the algorithm follows~\cite{hastie/etal:2009}. We use the indicator function $\mathbf{1}(A) =  1$ if $A$ is true and 0 otherwise. To train the predictor function for modelling $Y$, we specify hyperparameters $\delta, \lambda, d,$ and~$M$. The performance of the algorithm varies greatly with tree depth $d$ and number of boosting iterations $M$: tuning them carefully is important. Although the gradient boosting procedure aims at minimising the loss function through gradient descent, tuning is aimed at optimising generalisation to new observations. Minimising the training loss does not imply generalisation performance, which is why tuning should not be based on the training loss: an extrinsic metric, such as RMSE, MAE, or deviance, should be computed on new observations for various combinations of hyperparameters. The results are less sensitive to the subsampling percentage $\delta$ and the learning rate $\lambda$, as long as they are fixed to reasonable values. We detail below the role of each hyperparameter. 

In Step 1 of Algorithm~\ref{alg:gbm}, we initialise $f^0(\mathbf{x})$ at the constant $b$ which minimises the total loss for $\mathcal{D}$, the training set comprising $n$ records. Friedman \cite{Friedman:2002:StochasticGradient} introduces stochastic gradient boosting machines to prevent overfitting: in each iteration~$m \in~\{1, \ldots,  M\},$ we fit a decision tree on~$\mathcal{D}'$, a random subsample of size $\delta n$ from $\mathcal{D}$. For $i\in \mathcal{D}'$ in the $m$th iteration, we evaluate pseudo-residuals, defined as the gradient of the loss function evaluated at the previous iteration's prediction $f^{m-1}(\mathbf{x}_i)$:
\begin{align}
 \left.g_{i} = -\frac{\partial \mathcal{L}(y_i, b)}{\partial b}\right|_{b = f^{m - 1}(\mathbf{x}_i)}.
 \label{eq:gradient}
\end{align}
The pseudo-residuals indicate the direction towards minimal loss. In the learning phase, decision trees of depth $d$ are fitted to the pairs $\smash[b]{\{(\mathbf{x}_i, g_{i}):i \in \mathcal{D}'\}}$ with the mean squared error criterion --- regardless of the loss function $\mathcal{L}$ --- to partition the covariate space into regions with homogenous gradient values $g_i$. As mentioned in \cite{hastie/etal:2009}, the squared loss measures the closeness of the observations in terms of their gradient values, which already account for the shape of the data and the corresponding loss function. We emphasise that the target~$g_i$ in the decision trees reflects how ``well trained'' the $i$th observation is in iteration $m$: it changes with $m$. Yet, for simplicity of notation, we drop the index. Observations in undertrained regions of the covariate space have high absolute gradients, implying a large influence on the optimal tree splits due to the squared error criterion. Covariate space is partitioned according to gradient values by the decision trees in iterations $m\in\{1, \ldots, M\}$. Then, for terminal region $\smash[b]{j\in\{1, \ldots, J_m\}}$, we find the optimal step towards minimal loss, denoted~$\smash[b]{\hat{b}_{j}^m}$. We update the prediction $\smash{f^{m-1}}(\mathbf{x})$ for~$\mathbf{x} \in \smash{R^m_j}$ by adding~$\smash[b]{\hat{b}_{j}^m}$, shrunk by a learning rate~$\smash{\lambda \in (0, 1]}$. We use the \texttt{gbm3} implementation in \textsf{R} \cite{Hickey:2016}.

\begin{algorithm}[t]
    \SetKwFunction{isOddNumber}{isOddNumber}
    \SetKwInOut{KwIn}{Inputs}
    \KwIn{$\mathcal{L}, \varphi, \mathcal{D}, \delta$, $\lambda$, $d,$ and $M$.}
    \BlankLine
   Initialise $f^0(\mathbf{x}) =  \, \mbox{argmin}_b  \sum_{i = 1}^n \mathcal{L}(y_i, b).$ 
    \BlankLine
    \For{$m = 1$ \KwTo $M$}{
    \BlankLine
        Sample $\mathcal{D'}$, a subset of $\delta n$ observations without replacement from $\mathcal{D}.$        
    
    Compute $g_{i}$ with Eq.~\eqref{eq:gradient} for $i \in \mathcal{D}'.$
   
   Fit tree of depth $d$ to $\{(\mathbf{x}_i,g_{i}):i\in\mathcal{D}'\}$ to get terminal regions~$R^m_{1}, \ldots, R^m_{J_m}$ which greedily minimise the squared error.
   
   Compute, for $j \in \{1, \ldots, J_m\}$, $$\hat{b}^m_{j} =\underset{b}{\mbox{argmin}}{\sum\limits_{i \in\mathcal{D}': \mathbf{x}_i \in R^m_{j}} \mathcal{L}\{y_i, f^{m - 1}(\mathbf{x}_i) + b\}  }.$$ 
   
   Update $f^m(\mathbf{x}) = f^{m - 1}(\mathbf{x}) + \lambda \sum_{j = 1}^{J_m} \hat{b}^m_{j} \mathbf{1}{(\mathbf{x} \in R^m_{j})}.$}
    \BlankLine
    \textbf{Return~:}~Prediction function $f_{\mathrm{GBM}}(\mathbf{x}) = \varphi^{-1}\{f^M(\mathbf{x})\}$.
    \caption{Stochastic gradient boosting machine for decision trees}
    \label{alg:gbm}
\end{algorithm}

The learning rate, also known as the shrinkage parameter, helps convergence by preventing overly large steps and by avoiding local optima or saddle points. Shrinkage is a form of regularisation that mitigates over-specialisation, that is, when late iterations contribute effectively to the prediction of very few observations. Yet, Rashimi and Gilad-Bachrach \cite{Rashimi:2015} find that shrinkage does not overcome this issue. They propose the Dropout meets multiple Additive Regression Trees (DART) algorithm, inspired by the dropout approach used for neural networks \cite{srivastava/etal:2014}. They drop a proportion of the previous boosting iterations to evaluate the gradients, corrected by a scaling factor, instead of using~$f^{m-1}(\mathbf{x}_i)$ in Step 4 of Algorithm~\ref{alg:gbm}. DART involves hyperparameters to control the proportion and periodicity of the dropout. Rashimi and Gilad-Bachrach~\cite{Rashimi:2015} find that DART succeeds in tackling the over-specialisation problem in GBDT. We find that DART sometimes improves predictive performance over the use of a shrinkage parameter, at the cost of being computationally intensive.

It is found in \cite{Henckaerts/etal:2021} that GBMs improve performance over GLMs in insurance applications, most likely due to their flexibility and ability to automatically handle complex~($d>1$) non-linear effects. Yet, GBMs are computationally costly: each iteration involves an exhaustive search for CART splits and exact gradient evaluations. 

\subsubsection{Computational enhancements}
\label{subsec:computat}

Motivated by improving the scalability of the GBM algorithm, Chen and Guestrin \cite{Chen/etal.:2016:XGBoost} propose XGBoost. They leverage parallel programming and a simplified split-finding algorithm based on a quantile sketch of the gradients. XGBoost also optimises a second order Taylor approximation of the loss function with two regularisation parameters:~$\gamma$, the minimal loss function reduction required for a tree split, and $\phi$, the $\ell_2$ penalty on the tree predictions. The XGBoost objective in iteration $m \in \{1, \ldots, M\}$ is 
\begin{align*}
\sum_{i \in \mathcal{D}'} \left\{g_{i} h_m(\mathbf{x}_i) + t_{i} \frac{h_m(\mathbf{x}_i)^2}{2}  \right\} + \gamma J_m + \frac{1}{2} \phi \sum_{j = 1}^{J_m} \left(b^m_j\right)^2,
\end{align*}
where the gradient $g_i$ is defined in Eq.~\eqref{eq:gradient}, the Hessian of the loss function is
\begin{align}
\label{eq:hessian}
 \left.t_{i} = -\frac{\partial^2 \mathcal{L}(y_i, b)}{\partial b^2}\right|_{b = f^{m - 1}(\mathbf{x}_i)},
\end{align}
the decision tree prediction function is
$$
h_m(\mathbf{x}_i) = \sum_{j = 1}^{J_m} b^m_{j} \mathbf{1}{(\mathbf{x}_i \in R^m_{j})},
$$ 
and $b^m_j$ is the weight of leaf $j \in\{1,\ldots, J_m\}$. The $j$th optimal leaf weight is 
\begin{align}
\label{eq:leafweights}
\hat{b}^m_{j} = \frac{\sum\limits_{i\in \mathcal{D}':\mathbf{x}_i \in R^m_j } g_{i}}{\sum\limits_{i \in \mathcal{D}':\mathbf{x}_i \in R^m_j} t_{i} + \phi}.
\end{align}
Indeed, the Taylor approximation allows for a closed form computation of these tree predictions, speeding up the training even though we also compute the Hessian. We write the XGBoost procedure of \cite{Chen/etal.:2016:XGBoost} with our notation in Algorithm~\ref{alg:xgb}. As in random forest \citep{breiman:2001rf}, column subsampling reduces dimensionality when building the tree in Step~5 of Algorithm~\ref{alg:xgb}. We use the \textsf{R} implementation in the \texttt{xgboost} package~\citep{xgb}. 

\begin{algorithm}[t]
    \SetKwFunction{isOddNumber}{isOddNumber}
    \SetKwInOut{KwIn}{Inputs}
    \KwIn{$\mathcal{L}, \varphi, \mathcal{D}$, $\delta$, $\zeta$, $\lambda$, $d,$ $M$, $\gamma,$ and $\phi$.}
    \BlankLine
   Initialise $f^0(\mathbf{x}) =  \, \mbox{argmin}_b  \sum_{i = 1}^n \mathcal{L}(y_i, b).$ 
    \BlankLine
    \For{$m = 1$ \KwTo $M$}{
    \BlankLine
        Sample $\mathcal{D'}$, a subset of $\delta n$ observations without replacement from $\mathcal{D}.$        

    Compute $g_{i}$ and $t_{i}$ with Eqs~\eqref{eq:gradient} and~\eqref{eq:hessian} for $i \in \mathcal{D}'.$
   \BlankLine
   Fit tree of depth $d$ to $\{(\mathbf{x}_i,g_{i}):i\in\mathcal{D}'\}$ with the squared error loss, column subsampling $\zeta$, minimal loss reduction $\gamma$ and the simplified split-finding algorithm of \cite{Chen/etal.:2016:XGBoost} to get terminal regions $R^m_{1}, \ldots, R^m_{J_m}$.   
   \BlankLine
   Compute $\hat{b}^m_{j}$ using Eq.~\eqref{eq:leafweights} for $j \in \{1, \ldots, J_m\}$.
   
   Update $f^m(\mathbf{x}) = f^{m - 1}(\mathbf{x}) + \lambda \sum_{j = 1}^{J_m} \hat{b}^m_{j} \mathbf{1}{(\mathbf{x} \in R^m_{j})}.$}
    \BlankLine
    \textbf{Return~:}~Prediction function $f_{\mathrm{XGBoost}}(\mathbf{x}) = \varphi^{-1}\{f^M(\mathbf{x})\}$.
    \caption{XGBoost}
    \label{alg:xgb}
\end{algorithm}

LightGBM \citep{Ke/etal.:2017:LightGBM} and  CatBoost \citep{Prokhorenkova/Etal.:2018:Catboost} are enhancements of XGBoost. Both modify the tree growth strategy for Step 5 of Algorithm~\ref{alg:xgb}. In LightGBM, Ke et al. \cite{Ke/etal.:2017:LightGBM} aim to enhance computational efficiency in handling categorical features more succinctly, sophisticating the sampling step, and growing trees leaf-wise. CatBoost \cite{Prokhorenkova/Etal.:2018:Catboost} is designed for datasets with numerous, possibly imbalanced, categorical variables. It addresses GBDT algorithms' predictive shift, a bias proportional to $1/n$. In~\cite{So/Valdez:2024}, it is highlighted that CatBoost is suited for actuarial modelling because of the attention drawn to categorical features. A qualitative and quantitative comparative analysis of XGBoost, LightGBM, and CatBoost can be found in \cite{So:2024}. We experiment with the \textsf{R} implementations \texttt{lightgbm} \citep{lightgbm} and \texttt{catboost} \citep{catboost}. 

\smallskip
\begin{rem}
\label{rem:newton}
Second order Taylor approximation of the loss is called Newton boosting. According to Sigrist \cite{sigrist:2021}, it can improve predictive performance. Gradient and Newton boosting are used interchangeably in literature, which is criticised in \cite{sigrist:2021}. In Fig.~\ref{fig:implementations}, all the algorithms that are descendents of XGBoost use Newton steps.  
\end{rem}

\subsubsection{An interpretable alternative}
\label{sec:egbm}

GBDT algorithms with deep trees as weak learners are complex black boxes. Unlike GLMs, they do not output parameter estimates describing each covariate effect on the prediction. Also, high order interactions make it impossible for the human brain to conceptualise these effects. Thus, actuaries cannot directly explain a tariff structure output from a GBDT loss prediction model, as may be required by stakeholders or by regulation \citep[e.g.,][]{GDPR2016}. They must rely on---possibly misleading  \citep{xin/etal:2024-pdp}-- explanation tools. 

Transparent models can be obtained by taking a depth of $d=1$ in GBDT algorithms, because tree stumps split the covariate space according to only one feature at a time. To this end, Lou et al. \cite{lou:2013:GA2M, lou:2012:intelligibleCAR} build a generalised additive model (GAM) based on feature-wise effects (boosted tree stumps) and pairwise interactions: Generalised Additive Model plus Interactions. Following the \texttt{interpretml} \textsf{Python} package~\citep{Nori/etal:2019}, we refer to this method as EGBM.

EGBM follows a two-stage construction approach \citep{lou:2013:GA2M}, learning the main effects first and the interaction effects second. We initialise the prediction at constant $\beta_0$ in Step 1 of Algorithm~\ref{alg:egbm}. After $m~\in~\{1,\ldots, M\}$ iterations, the univariate function describing the main effect of feature $k\in\{1, \ldots, \eta\}$ is denoted $f^m_k$. It is learned with gradient boosting using only feature $k$ in a tree stump, cycling through the features. This is detailed in Steps 2--7 of Algorithm~\ref{alg:egbm}. For the $k$th main effect at iteration $m$, we evaluate gradients at the most recent prediction function, which is defined as
\begin{align*}
    f_k^{m*}(\mathbf{x}) &= \begin{cases}
         \beta_0 + \sum_{j=1}^{\eta} f^{m-1}_j(x_j), & \quad k = 1\\
         \beta_0 + \sum_{j=1}^{k-1} f^{m}_j(x_j) +  \sum_{j=k}^{\eta} f^{m-1}_j(x_j), & \quad k > 1.
    \end{cases}
\end{align*}
Note that the prediction function $f_k^{m*}$ takes the vector of covariates $\mathbf{x}$ as an argument while the main effect $f_k^m$ takes only the scalar $x_k.$ 

The FAST method of \cite{lou:2013:GA2M} allows for the selection of the set~$\mathcal{S}=\{\mathbf{s}_1,\ldots,\mathbf{s}_{n_{\mathrm{int}}}\}$ containing index couples of relevant two-way interactions. This method relies on the idea that an interaction should be included if it can notably reduce the loss of the main effect model. 
In the second construction stage (Steps 9--14 of Algorithm~\ref{alg:egbm}), the interaction effect~$\tilde{f}^m_{\ell}$ for pair $\mathbf{s}_\ell \in \mathcal{S}$ is learned by cycling through~$\ell\in\{1, \ldots, n_{\mathrm{int}}\}$. The interaction is built on top of the final prediction from the main effects using trees of depth $d =2$. In iteration~$m\in\{1, \ldots, M\}$, we note the most recent prediction function for the $\ell$th interaction effect between pair~$\mathbf{s}_\ell = (k, l)$, that is, between~$\mathbf{x}_{\mathbf{s}_\ell} = (x_k,x_l)$, as 
$$
 \tilde{f}_{\ell}^{m*}(\mathbf{x}) = \begin{cases}
        \beta_0 +  \sum_{j=1}^{\eta} f^{M}_j(x_j) + \sum_{j=1}^{n_{\mathrm{int}}} \tilde{f}_{j}^{m-1}(\mathbf{x}_{\mathbf{s}_j}), & \quad \ell = 1\\
         \beta_0 + \sum_{j=1}^{\eta} f^{M}_j(x_j) +  \sum_{j=1}^{\ell-1}\tilde{f}^{m}_{j}(\mathbf{x}_{\mathbf{s}_j}) + \sum_{j=\ell}^{n_{\mathrm{int}}} \tilde{f}^{m-1}_{j}(\mathbf{x}_{\mathbf{s}_j}), & \quad \ell > 1.
    \end{cases}
$$

Akin to a GAM, the equation of the EGBM prediction function is
\begin{equation*}
\varphi\left \{f_{\mathrm{EGBM}}(\mathbf{x})\right\} = \beta_0 + \sum_{k =1}^{\eta} f^M_k(x_{k}) + \sum_{\ell = 1}^{n_{\mathrm{int}}} \tilde{f}^M_{\ell}(\mathbf{x}_{\mathbf{s}_\ell}).
\end{equation*}
There is a predictor function for each variable and for each selected interaction; gradient boosting replaces the splines used in \cite{wood:2003}. These functions can be visualised and written explicitly as a formula or in a lookup table, making EGBM fully interpretable.

\begin{algorithm}[t!]
    \SetKwFunction{isOddNumber}{isOddNumber}
    \SetKwInOut{KwIn}{Inputs}
    \SetKwInOut{KwOut}{Output}
    \KwIn{$\mathcal{L}, \varphi, \mathcal{D}, \lambda, M,$ and $n_{\mathrm{int}}.$}
    \BlankLine
    Initialise $\beta_0 =\mbox{argmin}_b \sum_{i = 1}^n \mathcal{L}(y_i, b)$.
    \BlankLine
    \For{$m = 1$ \KwTo $M$}{
    \For{$k = 1$ \KwTo $\eta$}{
     Compute $g_{i}$ with Eq.~\eqref{eq:gradient} using $b = f_{k}^{m*}(\mathbf{x}_i)$ for $i \in \mathcal{D}$.

   Fit tree of depth $1$ to $\{(x_{k,i},g_{i}):i\in\mathcal{D}\}$ with the squared error loss to get regions $R^m_{1, k}$ and $R^m_{2, k}$.
   
   Compute, for $j \in \{1, 2\}$, $$\hat{b}^m_{j, k} =\underset{b}{\mbox{argmin}}{\sum\limits_{i \in\mathcal{D}: x_{k,i} \in R^m_{j, k}} \mathcal{L}\{y_i, f_k^{m*}(\mathbf{x}_i) + b\}  }.$$ 
   
   Update $f_k^m(x_k) = f_k^{m - 1}(x_k) + \lambda \sum_{j = 1}^{2} \hat{b}^m_{j, k} \mathbf{1}{(x_k \in R^m_{j, k})}.$
   
    }}
     Select $n_{\mathrm{int}}$ interactions with the FAST method \citep{lou:2013:GA2M} and get $\mathcal{S}$.
     
    \For{$m = 1$ \KwTo $M$}{
    \For{$\ell = 1$ to $n_{\mathrm{int}}$}{
    Compute $g_{i}$ with Eq.~\eqref{eq:gradient} using $b = \tilde{f}_{\ell}^{m*}(\mathbf{x}_i)$ for $i \in \mathcal{D}$.
   
   Fit tree of depth $2$ to $\{(\mathbf{x}_{\mathbf{s}_\ell,i},g_{i}):i\in\mathcal{D}\}$ with the squared error loss to get regions $\tilde{R}^m_{1, \ell}$ and $\tilde{R}^m_{2, \ell}$.
   
   Compute, for $j \in \{1, 2\}$, $$\tilde{b}^m_{j, \ell} =\underset{b}{\mbox{argmin}}{\sum\limits_{i \in\mathcal{D}: \mathbf{x}_{\mathbf{s}_{\ell},i} \in \tilde{R}^m_{j, \ell}} \mathcal{L}\{y_i, \tilde{f}_\ell^{m*}(\mathbf{x}_i) + b\}  }.$$ 
   
   Update $\tilde{f}_\ell^m(\mathbf{x}_{\mathbf{s}_\ell}) = \tilde{f}_\ell^{m-1}(\mathbf{x}_{\mathbf{s}_\ell}) + \lambda \sum_{j = 1}^{2} \tilde{b}^m_{j, \ell} \mathbf{1}{(\mathbf{x}_{\mathbf{s}_\ell} \in \tilde{R}^m_{j, \ell})}.$
    }}
    \BlankLine
    Aggregate $f^M(\mathbf{x})  = \beta_0  + \sum_{k=1}^{\eta} f_k^M(x_k) + \sum_{\ell=1}^{n_{\mathrm{int}}} \tilde{f}_\ell^M(\mathbf{x}_{\mathbf{s}_\ell}).$
    \BlankLine
    
    \textbf{Return : } Prediction function $f_{\mathrm{EGBM}}(\mathbf{x}) \leftarrow \varphi^{-1}\{f^M(\mathbf{x})\}.$
    \caption{Explainable gradient boosting machine}
    \label{alg:egbm}
\end{algorithm}

\subsection{Probabilistic gradient boosting for decision trees}
\label{sec:probalgos}

In Fig.~\ref{fig:implementations}, recently developed GBDT algorithms tackle the prediction of a probability distribution given $\mathbf{x}$. Most probabilistic GBDT approaches, such as XGBoostLSS, cyc-GBM, and NGBoost, aim to predict all the parameters of an assumed distribution. A different approach is proposed in \cite{sprangers2021probabilistic}: probabilistic gradient boosting machine (PGBM) outputs the mean and variance of the target variable for fixed~$\mathbf{x}$ by treating~$g_{i}, \, t_{i}$, and $\hat{b}_{j}^m$ from Eq.~\eqref{eq:leafweights} as random variables. Thus, PGBM is a nonparametric approach limited to location-scale distributional families, which excludes, e.g., the gamma. For this reason, we focus on multi-parametric probabilistic GBDT algorithms. These algorithms use $\kappa$ sequences of trees, one per element of $\mathbf{p} = (p_1, \ldots, p_{\kappa}),$ the vector parametrising the assumed distribution. In this section, we detail the algorithms of XGBoostLSS \citep{Marz:2019:xgboostlss}, cyc-GBM \citep{Delong/etal:2023}, and NGBoost \citep{Duan/etal.:2020:NGBoost}.

\subsubsection{XGBoostLSS}
XGBoostLSS \cite{Marz:2019:xgboostlss} extends the principles of gamboostLSS \citep{Hofner/Mayr:2016} to GBDT by joining the flexibility of GAMLSS \citep{rigby/stasinopoulos:2005} and the scalability of XGBoost \citep{Chen/etal.:2016:XGBoost}. It is designed as a ``wrapper around XGBoost'' \cite{Marz:2019:xgboostlss}. We use the negative log-likelihood of the assumed distribution as objective $\mathcal{L}$ and its first two partial derivatives w.r.t. each element of~$\mathbf{p}$.

\begin{algorithm}[h!]
    \SetKwFunction{isOddNumber}{isOddNumber}
    \SetKwInOut{KwIn}{Inputs}
    \SetKwInOut{KwOut}{Output}

    \KwIn{Negative log-likelihood $\mathcal{L},$ $\varphi_1, \ldots, \varphi_{\kappa}, \mathcal{D}, \delta, \zeta, \lambda, d, M,$ $\gamma$, $\phi,$ and $q_{max}$.}
    \BlankLine
    Initialise $q = 0$ and $\mathbf{f}_q^0(\mathbf{x})  = \{f_{1, q}^0(\mathbf{x}), \ldots, f_{\kappa, q}^0(\mathbf{x})\}=  \, \mbox{argmin}_{\mathbf{p}} \sum_{i = 1}^n \mathcal{L}(y_i, \mathbf{p}) \quad \forall q.$
    \BlankLine
    \For{$k = 1$ \KwTo $\kappa$}{
    \For{$m = 1$ \KwTo $M$}{
    \BlankLine  
        Sample $\mathcal{D'}$, a subset of $\delta n$ observations without replacement from $\mathcal{D}.$
        
        Compute $g_{i, k} = \frac{-\partial \mathcal{L}(y_i,\, \mathbf{p})}{\partial p_{k}}$ and $t_{i, k} = \frac{-\partial^2 \mathcal{L}(y_i,\, \mathbf{p})}{\partial p^2_{k}}$ with $\mathbf{p}=\hat{\mathbf{p}}_{k, q}^{m*}(\mathbf{x}_i)$ for~$i \in \mathcal{D}'.$
        \BlankLine
         Do Steps 5 and 6 of Algorithm~\ref{alg:xgb} using $g_{i, k}$ and $t_{i, k}$ in place of $g_i$ and $t_i$.
      
        Update $f_{k, q}^{m}(\mathbf{x}) = f_{k, q}^{m - 1}(\mathbf{x}) + \lambda \sum_{j = 1}^{J_{m, k}} \hat{b}^m_{j, k} \mathbf{1}{(\mathbf{x} \in R^m_{j, k})}.$
    }
    }
    \BlankLine
    \While{$\mathcal{L}$ has not converged and $q \leq q_{max}$}{
    Increment $q = q + 1$.
    
    \For{$k = 1$ \KwTo $\kappa$}{
     Update $f_{k,q}^M (\mathbf{x})$ by executing Steps 3 to 7.
    }}
    
    \textbf{Return :} Vector of prediction functions $$\mathbf{f}_{\mathrm{XGBoostLSS}}(\mathbf{x}) = \big[\varphi_1^{-1}\{f^M_{1,q} (\mathbf{x})\}, \ldots, \varphi_{\kappa}^{-1}\{f^{M}_{\kappa,q} (\mathbf{x})\}\big].$$

    \caption{XGBoostLSS}
    \label{alg:xgboostlss}
\end{algorithm}

Algorithm~\ref{alg:xgboostlss} presents XGBoostLSS, as introduced in \cite{Marz:2019:xgboostlss}, with our notation. We initialise each parameter estimate with MLE and allow for different link functions~$\varphi_1, \ldots, \varphi_{\kappa}$. A sequence of $M$ boosting iterations is made on each element~$k \in~\{1, \ldots, \kappa\}$ of $\mathbf{p}$. In each cycle, the XGBoost iterations update the~$k$th prediction function while treating the other parameters fixed. In cycle $q\in\{0,\ldots,q_{\mathrm{max}}\}$, when updating parameter $k$ at iteration $m \in\{1,\ldots, M\}$, we define the previous step vector of prediction functions as $\hat{\mathbf{p}}_{k, q}^{m*}(\mathbf{x}_i) = \{\hat{p}_{k, q, 1}^{m*}(\mathbf{x}), \ldots, \hat{p}_{k, q, \kappa}^{m*}(\mathbf{x})\}$ with $j$th element
\begin{align*}
    \hat{p}_{k, q, j}^{m*}(\mathbf{x}) = \begin{cases}
        f_{j, 0}^0(\mathbf{x}), &q = 0, \; k \neq j, \\
        f_{j, q}^{m-1}(\mathbf{x}), &q\geq0, \;k = j, \\
        f_{j, q-1}^{M}(\mathbf{x}), &q > 0, \;k < j, \\
        f_{j, q}^{M}(\mathbf{x}), &q>0, \;k > j.
    \end{cases}
\end{align*} 
For the first cycle, i.e., when $q=0$, the elements of $\mathbf{p}$ that are not currently being updated do not depend on $\mathbf{x}$: their initial values are used. In subsequent cycles, we evaluate at the most recent prediction function. We cycle until convergence of~$\mathcal{L}$ with pre-specified tolerance or until the process has been repeated $(q_{\mathrm{max}} + 1)$ times. The output is a vector of $\kappa$ prediction functions for distributional parameters given~$\mathbf{x}.$

\subsubsection{Cyclic gradient boosting machine}

A different approach to probabilistic GBDT is advocated in \cite{Delong/etal:2023}: cyc-GBM, a framework unifying multi-parametric boosting algorithms that use cyclic gradient descent. It includes the dual-parameter boosting of \cite{Meng/etal:2022}. In cyclic coordinate descent \citep[see, e.g., Section~8.8 of][]{luenberger/etal:2021}, distributional parameters are successively updated within each boosting iteration. Training time should be greater for cyc-GBM than for XGBoostLSS: cyclic gradient descent is more computationally intensive than Newton steps. The cyc-GBM algorithm relies on GBM, so optimisation is needed in each iteration.

\begin{algorithm}[h!]
    \SetKwFunction{isOddNumber}{isOddNumber}
    \SetKwInOut{KwIn}{Inputs}
    \SetKwInOut{KwOut}{Output}

    \KwIn{Negative log-likelihood $\mathcal{L}, \varphi_1, \ldots, \varphi_{\kappa},$ $\mathcal{D},  \lambda_1, \ldots, \lambda_{\kappa}, d_1, \ldots, d_{\kappa},$ and~$M_1, \ldots, M_{\kappa}$.}

    Initialise $\mathbf{f}^{0}(\mathbf{x}) = \{f_1^0(\mathbf{x}), \ldots, f_\kappa^0(\mathbf{x}) \}=  \, \mbox{argmin}_{\mathbf{p}} \sum_{i = 1}^n \mathcal{L}(y_i, \mathbf{p}) $ 
   
    \For{$m = 1$ \KwTo $M = \max(M_1,\ldots,M_\kappa)$}{
    \For{$k = 1$ \KwTo $\kappa$}{
    Fix $\mathbf{u}_{k},$ a unit (vertical) vector of length $\kappa$ with a 1 in the $k$th  position.
    
    \If{$m \leq M_{k}$}{
        Compute $ \left.g_{i, k} = -\frac{\partial \mathcal{L}(y_i, \mathbf{p})}{\partial p_{k}}\right|_{\mathbf{p} = \hat{\mathbf{p}}_{k}^{m*}(\mathbf{x}_i)}
$, for $i \in \mathcal{D}.$
    \BlankLine
    Fit tree of depth $d_{k}$ to $\{(\mathbf{x}_i,g_{i, k}):i\in\mathcal{D}\}$ with the squared error loss to get terminal regions~$R_{1, k}^m, \ldots, R_{J_{m, k},k}^m$. 
   \BlankLine
   For $j \in \{1, \ldots, J_{m, k}\},$ compute $$\hat{b}_{j, k}^{m} = \mbox{argmin}_b {\sum\limits_{i: \mathbf{x}_i \in R_{j, k}^m} \mathcal{L}\{y_i, \hat{\mathbf{p}}_{k}^{m*}(\mathbf{x}_i) + \mathbf{u}_{k} b\}  }.$$
   
   Update $f_{k}^m(\mathbf{x}) = f_{k}^{m - 1}(\mathbf{x}) + \lambda \sum_{j = 1}^{J_{m, k}} \hat{b}_{j, k}^m \mathbf{1}{(\mathbf{x} \in R_{j, k}^m)}
   .$
   }
   \BlankLine
   \textbf{else} Set $f_{k}^m (\mathbf{x}) = f_{k}^{m-1}(\mathbf{x}).$ 
    }}
    \BlankLine
    \textbf{Return : } Vector of prediction functions 
    $$\mathbf{f}_{\mathrm{cyc-GBM}}(\mathbf{x}) = \big[\varphi^{-1}_1\{f^{M}_1(\mathbf{x})\}, \ldots, \varphi^{-1}_\kappa\{f^{M}_{\kappa}(\mathbf{x})\}\big].$$
    \caption{Cyclic gradient boosting machine}
    \label{alg:cycgbm}
\end{algorithm}

Algorithm \ref{alg:cycgbm} presents the cyc-GBM algorithm of \cite{Delong/etal:2023} in our notation. When updating parameter $k \in\{1, \ldots, \kappa\}$ in iteration $m \in \{1, \ldots, M_k\}$, the previous step vector of prediction functions is
$\hat{\mathbf{p}}_k^{m*}(\mathbf{x}) = \{\hat{p}_{k, 1}^{m*}(\mathbf{x}), \ldots, \hat{p}_{k, \kappa}^{m*}(\mathbf{x}) \}$, with $j$th element 
$$
\hat{p}_{k, j}^{m*}(\mathbf{x}) = \begin{cases}
   f_{j}^{0}(\mathbf{x}), \quad & k \geq 1, m = 1, k \leq j, \\
    f_{j}^{m-1}(\mathbf{x}), \quad & k \geq 1, m > 1, k \leq j, \\
    f_{j}^{m}(\mathbf{x}), \quad & k  > 1, m\geq 1, k > j.
\end{cases}
$$

The cyc-GBM algorithm \cite{Delong/etal:2023} is designed to enhance flexibility in multi-parametric GBDT. Different tree depth, learning rate, and number of trees per distributional parameter are allowed in cyc-GBM. This feature may prevent overfitting on some distributional parameters while others are still undertrained. For enhanced explainability, we can set~$d_k = 0$ for some $k$ if, e.g., we do not want a parameter to vary with~$\mathbf{x}.$

\subsubsection{Natural gradient boosting}

In XGBoostLSS and cyc-GBM, the change indicated by the computed gradients~$g_{i, k}$ does not necessarily correspond to an equivalent change in probability space. As explained in \cite{Duan/etal.:2020:NGBoost}, ``[t]he problem is that `distance' between two parameter values does not correspond to an appropriate `distance' between the distributions that those parameters identify.'' This motivates the use of the natural gradient in NGBoost~\cite{Duan/etal.:2020:NGBoost}.

The classical gradient $\nabla\mathcal{L}(y, \mathbf{p})$ lacks the property of invariance under reparametrisation. An invariant alternative is the generalised natural gradient, defined as
$$
\widetilde{\nabla}\mathcal{L}(y, \mathbf{p}) \propto \mathcal{I}_{\mathcal{L}}^{-1}(\mathbf{p}) \nabla\mathcal{L}(y, \mathbf{p}),
$$
where $\mathcal{I}_{\mathcal{L}}$ is the Fisher information matrix $\E[\nabla\mathcal{L}(Y, \mathbf{p}) \nabla\mathcal{L}(Y, \mathbf{p})^{\top}]$  when the loss function is the negative log-likelihood. The disadvantage of the natural gradient is that its evaluation requires the inversion of a $\kappa \times \kappa$ matrix in each iteration $m\in\{1, \ldots, M\}$.

\smallskip
\begin{rem}
     The natural gradient can be generalised to the use of any proper scoring rule as a loss function. We refer to \cite{Duan/etal.:2020:NGBoost} for the technical details.
\end{rem}
\smallskip

Algorithm \ref{alg:ngb} is the NGBoost procedure of \cite{Duan/etal.:2020:NGBoost} in our notation. This algorithm is very close to a vectorised version of Algorithm~\ref{alg:gbm}. In each iteration of NGBoost, $\kappa$ trees are built in Step 4. The vector of tree predictions $\mathbf{h}^m(\mathbf{x})$ is multiplied by a scalar~$\hat{\rho}^m,$ optimised globally in Step 5, before being added to the vector of prediction functions in the update of Step 6. Interestingly, the prediction of the decision tree participates directly to the NGBoost prediction function: in other GBDT algorithms, the update depends on the tree only through terminal regions. Also, unlike in XGBoostLSS and cyc-GBM, all parameter prediction functions are updated simultaneously. As hinted in \cite{Duan/etal.:2020:NGBoost}, NGBoost could predict multivariate responses with a joint log-likelihood. We use the \textsf{Python} implementation \texttt{ngboost} \cite{Duan/etal.:2020:NGBoost} in this study.

\begin{algorithm}
    \SetKwFunction{isOddNumber}{isOddNumber}
    \SetKwInOut{KwIn}{Inputs}
    \SetKwInOut{KwOut}{Output}

    \KwIn{Negative log-likelihood $\mathcal{L},$ $\varphi_1, \ldots, \varphi_{\kappa}$,  $\mathcal{D}, \lambda, d,$ and $M$.}
    \BlankLine
    Initialise $\mathbf{f}^0(\mathbf{x}) =  \, \mbox{argmin}_{\mathbf{p}} \sum_{i = 1}^n \mathcal{L}(y_i, \mathbf{p}).$
    \BlankLine
    \For{$m = 1$ \KwTo $M$}{
    \BlankLine
    Compute $\mathbf{g}_{i} = \Tilde{\nabla} \mathcal{L}(y, \mathbf{p}) \big|_{\mathbf{p} = \mathbf{f}^{m - 1}(\mathbf{x}_i)}
   $, for $i \in \mathcal{D}.$
   
    \textbf{for} $k = 1$ \KwTo $\kappa$, fit tree  of depth $d$ to $\{(\mathbf{x}_i,g_{i, k}):i\in\mathcal{D}\}$ with the squared error loss to get predictor $ h_{k}^m(\mathbf{x})$.
  
   \BlankLine
   
   Optimise, by denoting $\mathbf{h}^m(\mathbf{x})=\{h_{1}^m(\mathbf{x}),\ldots, h_{\kappa}^m(\mathbf{x})\}$, the scalar 
   $$\hat{\rho}^m = \mbox{argmin}_\rho \sum\limits_{i=1}^n \mathcal{L}\{y_i, \mathbf{f}^{m - 1}(\mathbf{x}_i) + \rho \mathbf{h}^m(\mathbf{x}_i)\}.$$ 
   
   Update $\mathbf{f}^m(\mathbf{x}) = \mathbf{f}^{m - 1}(\mathbf{x}) + \lambda \hat{\rho}^m \mathbf{h}^m(\mathbf{x}).$
    \BlankLine
    }
    \BlankLine
    \textbf{Return : } Vector of prediction functions $$\mathbf{f}_{\mathrm{NGBoost}}(\mathbf{x}) = \big[\varphi_1^{-1}\{f_{1}^M(\mathbf{x})\}, \ldots,  \varphi_{\kappa}^{-1}\{f_{\kappa}^M(\mathbf{x})\}\big].$$
    \caption{Natural gradient boosting}
    \label{alg:ngb}
\end{algorithm}

\subsection{Algorithmic hybridisation}
\label{sec:hybrid}

The algorithms presented in this section correspond to the nodes in Fig.~\ref{fig:implementations}. To the best of our knowledge, this is a comprehensive survey of the existing point and multi-parametric probabilistic GBDT algorithms in literature. However, creating hybrid algorithms is possible. For example, the DART procedure can always replace shrinkage.

Examples of algorithmic hybrids are CatBoostLSS \citep{marz:2020} and LightGBMLSS \citep{marz:2023}  which replace XGBoost by CatBoost or LightGBM in Algorithm~\ref{alg:xgboostlss}. The DART algorithm is used in~\cite{Meng/etal:2022} to perform dual-parameter boosting for claim frequency modelling. We use the DART procedure along with XGBoostLSS instead of shrinkage, which we denote XGBoostLSSd. The natural gradient, as in NGBoost, is coupled with LightGBM to predict the cost of novel constructions in \cite{Chakraborty/etal:2020}. Finally, any GBDT prediction algorithm can leverage shallow trees and selected two-way interactions as in EGBM to enhance model interpretability. The possibilities are numerous; we focus on comparing the native algorithms in our numerical experiment. 

\section{Methods}
\label{sec:methodology}

Our objective is to compare the performance of the algorithms presented in Section~\ref{sec:survey} on actuarial data. We also include for reference a vanilla GLM without interaction and a GAMLSS with cubic splines on the main effects only. Distributional assumptions for claim frequency and severity are set in Section~\ref{sec:context} with a special emphasis on the treatment of varying exposure-to-risk. Then, we present the evaluation framework in Section~\ref{sec:comparisonpoints}. Appendix~\ref{app:tuning} details training and tuning strategies. Hyperparameter tuning is based on an extrinsic metric computed on a separate validation set to enhance generalisation. In the same vein, some mitigation strategies, including shrinkage, cross-validation, and regularisation, can be employed to cope with overfitting, which can arise in GBDT models.

\subsection{Distributional assumptions for claim frequency and severity}
\label{sec:context}

We consider the modelling of insurance claim frequency, with the Poisson and negative binomial distributions, and severity, with the lognormal and gamma distributions.

Within an insurance portfolio, each policy $i\in\{1, \ldots, n\}$ is observed for a given duration $e_i$. Claim frequency modelling should take into account varying exposure-to-risk. In a Poisson GLM with log link, this is classically handled with an offset term~$\ln e_i$. We adopt this idea in the GBDT framework: the initial prediction for the claim frequency of policy $i$ is
$\ln e_i + f^0 (\mathbf{x}_i),$ which is refined through gradient boosting. The offset~$\ln e_i$ is introduced in GBDT implementations with the argument called \emph{initial\_score}, \emph{baseline} or \emph{base score} depending on the package.

Because insurance claim data is often overdispersed, we also consider the negative binomial distribution. The NB2 distribution with probability mass function 
\begin{equation*}
f(y_i) = \frac{\Gamma(\phi_i + y_i)}{\Gamma(\phi_i) \Gamma(y_i + 1)}\left(\frac{\mu_i}{\mu_i + \phi_i} \right)^{y_i} \left(\frac{\phi_i}{\mu_i + \phi_i} \right)^{\phi_i}, \quad y_i \in \mathbb{N} \text{ for } i \in \{1, \ldots, n\},
\end{equation*} 
is conveniently parametrised by a location parameter $\mu >0$ and a dispersion parameter~$\phi > 0$. This parametrisation allows to treat varying exposure-to-risk as an offset for parameter $\mu_i$ while treating $\phi_i$ without the exposure-to-risk. The principles of Section~\ref{sec:frompointtoprob} apply: dispersion parameter $\phi_i$ can be estimated either individually through a multi-parametric GBDT algorithm or globally~($\phi_i := \phi \; \forall i$) with MLE.

For the lognormal assumption on claim severity, we use $\ln Y$ as the target with a normal distribution and the squared error loss function. For convenience, we parametrise the gamma distribution with mean $\mu > 0$ and shape $\alpha > 0$.

\smallskip
\begin{rem}
Approaches that directly model the loss cost with a Tweedie distribution are implemented, for example, with GBM in \cite{Denuit/etal:2021}. The Tweedie loss function is readily available in packages \texttt{gbm3}, \texttt{xgboost}, \texttt{lightgbm}, \texttt{catboost}, \texttt{interpretml}, and \texttt{pgbm}.
\end{rem}

\subsection{Performance evaluation framework}
\label{sec:comparisonpoints}

We measure computational efficiency by the training time in seconds on a personal laptop computer (IntelCore i7-1195G7 @ 2.90 GHz CPU) for a fixed combination of hyperparameters ($M = 1000,$ $d = 5$, $\delta = 0.75$, and $\lambda = 0.01$). This way, we compare the difference induced by the operations of each algorithm without regards to tuning. Higher computational time in more complex models is expected, but it does not imply that more complex models should not be used. However, insurers work with heavy datasets, and because society calls for digital sobriety, the computational cost is an environmental cost and must be considered along with predictive performance.

As evoked in Section~\ref{sec:frompointtoprob}, we use the deviance to measure predictive performance. With negative log-likelihood $\mathcal{L},$ the deviance of observation $i \in \{1, \ldots, n\}$ is
$$
D\{y_i, f_{\mathrm{GBDT}}(\mathbf{x}_i)\} = 2 \left[-\mathcal{L}(y_i, y_i) + \mathcal{L}\{y_i, f_{\mathrm{GBDT}}(\mathbf{x}_i)\}\right].
$$
Although the deviance measures the predictive performance of~$f_{\mathrm{GBDT}}$, its absolute value does not have an inherent meaning or scale. Inspired by~\cite{So:2024}, we thus present an alternative predictive performance metric named McFadden's pseudo-$R^2$, which measures the improvement in predictive performance compared to the constant $\bar{y}$ model \citep{mcfadden:1974}. For a test set of $n_{test}$ observations, the expression is
\begin{align}
R^2 &= 1 - \frac{\sum_{i = 1}^{n_{test}}D\{y_i, f_{\mathrm{GBDT}}(\mathbf{x}_i)\}}{\sum_{i = 1}^{n_{test}}D(y_i, \bar{y})}. \label{eq:pseudor2}
\end{align} A model is better if its McFadden's pseudo-$R^2$ is closer to one.

We adopt the consistent scoring function framework proposed in \cite{ehm/etal:2016} to compare the predictive performance of the algorithms on the test sets. While Eq.~\eqref{eq:pseudor2} relies on a chosen loss function (the deviance), analysing predictive dominance allows to evaluate if this choice dictates the verdict about predictive performance.  We draw Murphy diagrams, as in \cite{fissler/etal:2023,holvoet/etal:2025}, to evaluate predictive dominance, comparing models based on the elementary scoring function
{\small$$
S_{\nu}\{f_{\mathrm{GBDT}}, \mathcal{D}_{\mathrm{test}}\} = \frac{1}{n_{\mathrm{test}}} \sum_{i=1}^{n_{\mathrm{test}}} |\nu - y_i| \mathbf{1}[\min\{f_{\mathrm{GBDT}}(\mathbf{x}_i), y_i\} \leq \nu < \max\{f_{\mathrm{GBDT}}(\mathbf{x}_i), y_i\}],
$$}where parameter $\nu \in \mathbb{R}$ can be interpreted as an oracle prediction. As seen in Section 3.4 of \cite{ehm/etal:2016}, predictive dominance of model $f'$ on model $f$ arises if and only if $S_{\nu}\{f', \mathcal{D}_{\mathrm{test}}\} \leq S_{\nu}\{f, \mathcal{D}_{\mathrm{test}}\}$ holds for all $\nu \in \mathbb{R}$, a rather stringent requirement.

Model adequacy informs on how well an estimated assumed distribution fits the data. In particular, a model is inadequate if some observations are highly unlikely given the estimated parameters. We quantify model adequacy with proper scoring rules, functions that assign a score to a probabilistic prediction  given the corresponding observation.  Smaller values of proper scoring rules indicate better model adequacy~\citep[see, e.g,][]{gneiting:2005:calibratedprobabilisticforecasting}. For an estimated cumulative distribution function $F_{\hat{\mathbf{p}}}$, the continuous ranked probability score (CRPS) is given by
$$
CRPS(F_{\hat{\mathbf{p}}}, y_i) = \int_{-\infty}^{\infty}\left\{F_{\hat{\mathbf{p}}}(z) - \mathbf{1}(y_i \leq z) \right\}^2 dz.
$$
The \textsf{R} package \texttt{scoringRules} \cite{jordan/collab:2017:evaluatingScorignRules} implements the CRPS. With frequency data, we use double probability integral transform (DPIT) residuals \cite{yang:2024}, which should follow a uniform distribution when the model is adequate.

\begin{figure}[t]
    \centering
    \includegraphics[width = 0.8\textwidth]{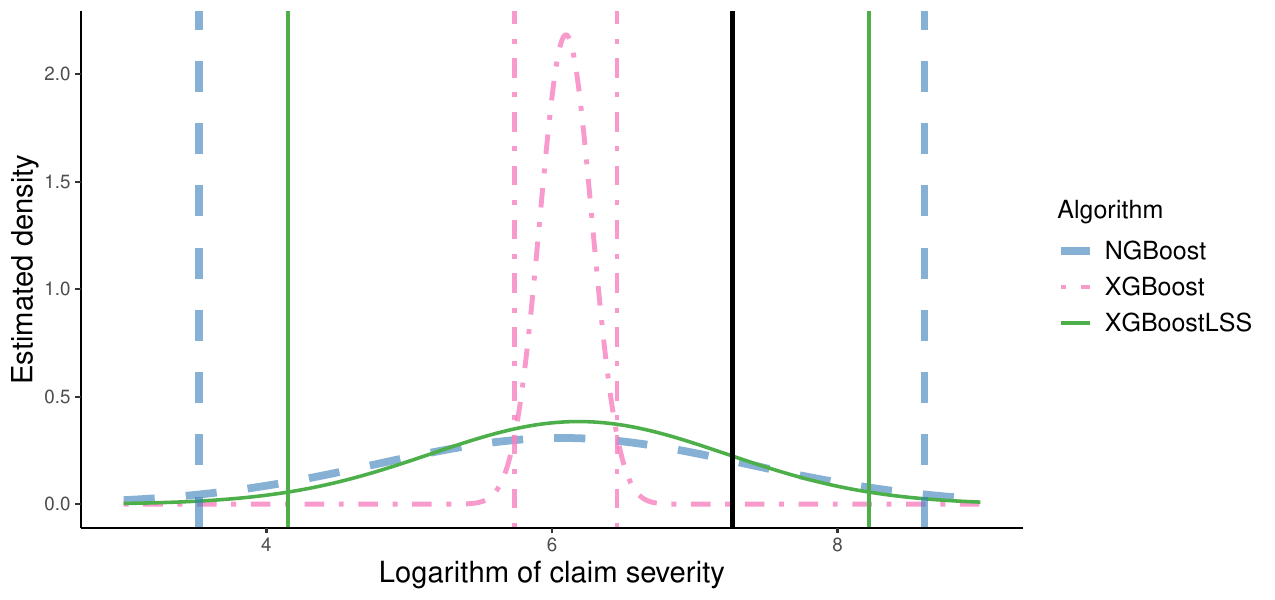}
    \caption{Predicted lognormal densities given $\mathbf{x}_1$ by three algorithms (color and line type), corresponding 95\% confidence intervals, and target $y_1$ (black bold vertical line).}
    \label{fig:explic}
\end{figure}

To measure the adequacy of continuous distributions to severity data, we also consider the coverage of confidence intervals (CIs) on the test set. The probabilistic forecast of an observation leads to an estimated distribution function from which we can compute CIs. The proportion of observations in the test set that fall inside their corresponding predicted CI is the coverage. In an adequate model, CI coverage is near CI level. In Fig.~\ref{fig:explic}, the observed response drawn with a black vertical line falls within the 95\% CIs predicted by NGBoost and XGBoostLSS (blue and green vertical lines). However, the XGBoost model (pink) does not fit well this observation. Note that we use confidence intervals to analyse model adequacy. These are not prediction intervals, which would lead to an analysis of prediction uncertainty. Here, we compare the competing algorithms under a given distributional assumption. This differs from the usual setup, where distributions obtained with a given algorithm compete. 

From an actuarial standpoint, portfolio balance and calibration are important to ensure that a sufficient amount of premium is collected. Our evaluation framework addresses portfolio balance by measuring the relative difference between the predicted and the observed response on the test set. We visually assess calibration by plotting the observed response against its predicted counterpart over binned prediction ranges.

Calibration is numerically evaluated by performing the statistical test for auto-calibration based on the difference between Lorenz curves and concentration curves proposed by \cite{Denuit/etal:2024}. We perform the statistical test with the test set that was not seen during training or tuning~$\mathcal{D}_{test} = \{(\mathbf{x}_i, y_i): i = 1, \ldots, n_{test}\}$. Inspired by the notation of \cite{Denuit/etal:2024}, we have the concentration curve~$CC\{\E(Y|\mathbf{X} = \mathbf{x}), f_{\textrm{GBDT}}(\mathbf{x}); \beta \}$   and the Lorenz curve~$LC\{f_{\textrm{GBDT}}(\mathbf{x}); \beta\}$ for every level $\beta \in [0,1].$ According to Proposition 2.1 of \cite{Denuit/etal:2024}, the rebalanced predictions are auto-calibrated if and only if
$$
CC\{\E(Y|\mathbf{X} = \mathbf{x}), f_{\textrm{GBDT}}(\mathbf{x}); \beta \} = LC\{f_{\textrm{GBDT}}(\mathbf{x}); \beta\} \quad \forall \; \beta \in [0, 1],
$$ 
which is the null hypothesis of the statistical test for auto-calibration. We test against the two-sided alternative at level $\alpha = 0.01$ by following the procedure in Section 4 of \cite{Denuit/etal:2024} with $B=1000$. It is trivial to rebalance the GBDT predictions based on the training data, which we do in the evaluation of portfolio balance and auto-calibration.

\section{Results}
\label{sec:results}

Performances of the point and probabilistic GBDT algorithms in Fig.~\ref{fig:implementations} are compared within the same experimental setup on various publicly available datasets. 

The characteristics of the publicly available datasets used for frequency modelling are presented in Table~\ref{tab:freqdata}. All of them are non-synthetic and include a column for the exposure $e_i$. The diversity in sample size, number of features, number of categorical variables, and maximum number of levels for categorical variables allows to assess the influence of these characteristics on performances.   In Table~\ref{tab:freqdata}, we list which source inspired the preprocessing. Note that, for pg15training, we only use the 2009 cohort for claim frequency modelling. We present the datasets for severity modelling in Table~\ref{tab:sevdata}.  Most datasets are from \texttt{CASDatasets}~\citep{CASdatasets}; BelgianMTPL is introduced in \cite{Denuit/Lang:2004}, Emcien can be found in \cite{emcien}, and the synthetic WorkComp data in~\cite{compensation}.

\begin{table}[h]
\caption{Characteristics of the claim frequency datasets.}
\label{tab:freqdata}
\begin{tabular}{lcrrrr}
\toprule
Name & Preprocessing & Sample size & \begin{tabular}{c}
     \# of \\
     features
\end{tabular}& \begin{tabular}{c}
\# of cat.\\ variables \end{tabular} & \begin{tabular}{c}
Max. \# of  levels \\ for cat. variables \end{tabular}\\ \midrule
   freMTPL & {\small \cite{So:2024}}  & 678~013~~ & 9~~~~~~~ & 4~~~~~~~~~ & 21~~~~~~~~~~~~ \\
     freMPL & {\small \cite{henckaerts/etal:2022}} &165~200~~ & 9~~~~~~~ & 6~~~~~~~~~ & 46~~~~~~~~~~~~ \\
   BelgianMTPL  &  {\small \cite{Henckaerts/etal:2021}} & 163~212~~ & 11~~~~~~~ & 5~~~~~~~~~ & 3~~~~~~~~~~~~  \\
    swauto  &  {\small \cite{ohlsson/Johanson:2010}} & 62~436~~ & 6~~~~~~~ & 4~~~~~~~~~ & 7~~~~~~~~~~~~ \\
pg15training9 & {\small \cite{xin/etal:2024}} & 50~000~~ & 13~~~~~~~ & 8~~~~~~~~~ & 471~~~~~~~~~~~~ \\\bottomrule
\end{tabular}
\end{table}

\begin{table}[h]
\caption{Characteristics of the claim severity datasets.}
\label{tab:sevdata}
\begin{tabular}{lcrrrr}
\toprule
Name &  Preprocessing & Sample size & \begin{tabular}{c}
     \# of \\
     features
\end{tabular}& \begin{tabular}{c}
\# of cat.\\ variables \end{tabular} & \begin{tabular}{c}
Max. \# of  levels \\ for cat. variables \end{tabular} \\ \midrule
    WorkComp & &  48~703~~ & 21~~~~~~~ & 3~~~~~~~~~ & 3~~~~~~~~~~~~ \\
    freMTPL & \cite{So:2024} & 21~611~~ & 9~~~~~~~ & 4~~~~~~~~~ & 21~~~~~~~~~~~~  \\
    BelgianMTPL  & \cite{Henckaerts/etal:2021} & 17~910~~ & 11~~~~~~~ & 5~~~~~~~~~ & 3~~~~~~~~~~~~  \\
    pg15training  & \cite{xin/etal:2024} & 12~256~~ & 14~~~~~~~ & 9~~~~~~~~~ & 471~~~~~~~~~~~~~\\ 
    Emcien & & 9~134~~ & 17~~~~~~~ & 12~~~~~~~~~ & 9~~~~~~~~~~~~ \\
    \bottomrule
\end{tabular}
\end{table}

We preprocess data and compute results with \textsf{R}. However, the \textsf{Python} implementations of EGBM, XGBoostLSS, and cyc-GBM are used because there is no equivalent in~\textsf{R}. We present results on computational efficiency in Section~\ref{sec:efficiency} and predictive performance in Section~\ref{sec:predperformance}. In Section~\ref{sec:adequacy}, we analyse model adequacy.

\subsection{Computational efficiency}
\label{sec:efficiency}

The results on computational efficiency for the Poisson and NB2 frequency models are presented in Tables~\ref{tab:freqefficiency} and~\ref{tab:NBefficiency}, respectively.  Point NB2 predictions are obtained by estimating a global dispersion parameter out of the mean parameter $\mu(\mathbf{x})$ output in Poisson models, which is why point algorithms are not present in Table~\ref{tab:NBefficiency}. The horizontal line delimits the point and probabilistic  algorithms; we provide the average ranking over all datasets within each type of algorithm in the last column. LightGBM and XGBoostLSS stand out in their category as the most computationally efficient. Other key takeaways of Tables~\ref{tab:freqefficiency}--\ref{tab:NBefficiency} are:
\begin{itemize}
    \item The major gap between training time for XGBoostLSS with shrinkage and with DART illustrates that the latter is time-consuming.
    \item Predicting the dispersion parameter in addition to the location parameter for the NB2 distribution more than doubles the training time in most of the cases.
    \item Sample size (decreasing from left to right in the tables) and the number of levels in categorical features impact training time. The comparison between freMPL and BelgianMTPL illustrates the influence of the cardinality of categorical variables.
    \item  XGBoost and LightGBM have smaller training times than GBM, as expected.
\end{itemize}

\begin{table}[h]
\centering
\caption{Training time in seconds for Poisson frequency models.}
\label{tab:freqefficiency}
\begin{tabular}{lrrrrrr}
\toprule
\multirow{2}{*}{Model} & \multicolumn{5}{c}{Dataset} & \multirow{2}{*}{Avg. rank} \\\cmidrule{2-6}
 & freMTPL & freMPL & BelgianMTPL & swauto & pg15training & \\ \midrule
    LightGBM  & 20 & 6 & 7 & 3 & 6 & 1.0\\
    XGBoost & 1~120 & 38 & 34 & 22 & 13& 2.4\\
    EGBM  & 1~015 & 481 & 82 & 15 & 49 & 3.4\\
    GBM  & 3~829 & 50 & 38 & 24 & 17 & 3.6\\
    CatBoost  & 1~780 & 106 & 84 & 34 & 51 & 4.6\\\midrule
    XGBoostLSS  & 1~153 & 38 & 35 & 24 & 15 & 1.0\\
    NGBoost  & 4~653 & 361 & 223 & 65 & 145 & 2.0\\
    cyc-GBM  & 9~313 & 452 & 349 & 136 & 238 & 3.2\\
    XGBoostLSSd  & 10~341 & 757 & 264 & 414 & 377 & 3.8\\
\bottomrule
\end{tabular}
\end{table}

\begin{table}[h]
\centering
\caption{Training time in seconds for NB2 frequency models.}
\label{tab:NBefficiency}
\begin{tabular}{lrrrrrr}
\toprule
\multirow{2}{*}{Model} & \multicolumn{5}{c}{Dataset} & \multirow{2}{*}{Avg. rank} \\\cmidrule{2-6}
& freMTPL & freMPL & BelgianMTPL & swauto & pg15training &\\ \midrule
    XGBoostLSS  & 3~248 & 112 & 60 & 75 & 55 & 1.0\\
    NGBoost  & 8~532 & 1~100 & 424 & 101 & 252 & 2.0\\
    cyc-GBM  & 16~034 & 2~646 & 948 & 705 & 863 & 3.4\\
    XGBoostLSSd  & 17~360 & 2~833 & 514 & 805 & 731 & 3.6\\
\bottomrule
\end{tabular}
\end{table}

Table~\ref{tab:sevefficiency} in Appendix~\ref{app:addres} displays the training time for lognormal models and leads to similar takeaways. Results for PGBM and XGBoost are similar because the same algorithm is used; training times are slightly higher in PGBM due to the estimation of variances in each terminal tree node. Similar conclusions were obtained with the gamma distribution (not shown).

\subsection{Predictive performance}
\label{sec:predperformance}

Tables \ref{tab:perfpoisson}--\ref{tab:sevgamma} show McFadden's $R^2$ for the Poisson, NB2, lognormal, and gamma models, respectively. The last column is the average ranking over all datasets and models. Algorithms are ordered following the computational efficiency average ranks. Note that, in a Poisson model, point and probabilistic predictions are equivalent.

For a given dataset, the magnitude of the $R^2$ is broadly the same for all algorithms: probabilistic forecasting does not hinder much predictive performance. This is also seen in the average ranking.  We see that GBDT models can improve performance over GLM and GAMLSS.  In terms of average rank, CatBoost wins, whereas cyc-GBM looses, in three of the four tables. The specific design of CatBoost seems advantageous, especially for datasets with high cardinality categorical features, commonly seen in actuarial pricing. Interestingly, EGBM has a slightly better performance than GBM while being fully interpretable. Also, XGBoostLSS performs better with DART rather than shrinkage on all frequency datasets but swauto at the cost of an increased training time (see Section~\ref{sec:efficiency}). This is not observed in severity models. The relative ordering of LightGBM, XGBoost, and CatBoost with freMTPL on Poisson boosting is consistent with results in \cite{So:2024}.

\begin{table}[h!]
\centering
\caption{McFadden's pseudo-$R^2$ of test Poisson deviance (in \%).}
\label{tab:perfpoisson}
\begin{tabular}{lrrrrrr}
\toprule
\multirow{2}{*}{Model} & \multicolumn{5}{c}{Dataset} & \multirow{2}{*}{Avg. rank} \\\cmidrule{2-6}
 & freMTPL & freMPL & BelgianMTPL & swauto & pg15training & \\ \midrule
    GLM & 20.16 & 59.51 & 3.78 & 24.47 & 21.86 & 10.4\\
    GAMLSS & 20.78 & 66.49 & 4.37 & 26.22 & 22.80 & 6.2\\\midrule
    LightGBM  & \textbf{24.92} & 65.86 & 4.52 & \textbf{26.31} & 22.79 & 4.0 \\
    XGBoost & 24.87 & 66.68 & 4.45 & 24.81 & 22.42 & 6.0\\
    EGBM  &  23.92 & 65.99 & \textbf{4.64} & 25.97 & 24.95 & 3.8\\
    GBM  & 23.85 & 65.14 & 4.63 & 26.03 & 24.11 & 4.8\\
    CatBoost  & 24.91 & 65.67 & 4.56 & 26.25 & \textbf{24.99} & \textbf{3.0}\\ \midrule
    XGBoostLSS  & 24.84 & 66.77 & 4.39 & 24.92 & 21.39 & 6.2\\
    NGBoost  & 22.87 & 62.42 & 4.31 & 24.56 & 23.61 & 7.8\\
    cyc-GBM  & 16.85 & 60.16 & 3.93 & 21.49 & 22.88 & 9.6\\
    XGBoostLSSd  & 24.90 & \textbf{66.94} & 4.48 & 24.83 & 22.98 & 4.2\\
    \bottomrule
\end{tabular}
\end{table}

\begin{table}[h!]
\centering
\caption{McFadden's pseudo-$R^2$ of test negative binomial deviance (in \%).}
\label{tab:perfnb}
\begin{tabular}{lrrrrrr}
\toprule
\multirow{2}{*}{Model} & \multicolumn{5}{c}{Dataset} & \multirow{2}{*}{Avg. rank} \\\cmidrule{2-6}
& freMTPL & freMPL & BelgianMTPL & swauto & pg15training & \\ \midrule
    GLM & 22.37 & 60.43 & 1.68 & 23.50 & 10.78 & 4.4\\
    GAMLSS & 19.88 & 62.44 & 1.70 & 24.04& 10.86 & 3.6\\\midrule
    XGBoostLSS & 20.84 & 61.29 & 1.71 & \textbf{25.02} & 11.80 & 3.0 \\
    NGBoost & \textbf{22.56} & 61.84 & 1.48 & 24.98 & 12.61 & 2.8\\
    cyc-GBM  &  12.49 & 54.72  & \textbf{3.62} & 21.91 & 10.82 & 4.8\\
    XGBoostLSSd & 22.16 & \textbf{62.77} &  2.69 & 23.11&   \textbf{13.30} & \textbf{2.4}\\
    \bottomrule
\end{tabular}

\end{table}

\begin{table}[h!]
 \centering
\caption{McFadden's pseudo-$R^2$ on the test set based on RMSE for the lognormal models (in \%).}
\label{tab:sevlognorm}
\begin{tabular}{lrrrrrr} \toprule
\multirow{2}{*}{Model} & \multicolumn{5}{c}{Dataset} & \multirow{2}{*}{Avg. rank} \\\cmidrule{2-6}
& WorkComp & freMTPL & BelgianMTPL & pg15training & Emcien & \\ \midrule 
GLM & 25.41 & 0.56 & 0.27 & 1.09 & 48.85 & 9.0\\
GAMLSS & 25.87& 0.63 & 0.38 & 0.64& 49.37 & 9.2\\\midrule
LightGBM & \textbf{29.83} & 0.71 & 0.44 & 0.34 & 49.07 & 6.4\\ 
XGBoost & 29.03 & 0.81 & \textbf{0.51} & 1.22 & 48.89 & 4.0\\
EGBM & 29.59 & 0.76 & 0.45 & 1.52 & 48.85 & 4.8\\
GBM & 26.27 & 0.78 & 0.48 & 0.60 & 48.89 & 6.2\\
CatBoost & 24.11 & \textbf{0.83} & 0.50 & \textbf{1.52} & \textbf{50.89} & \textbf{3.2}\\\midrule 
PGBM & 29.01 & 0.80 & 0.50 & 1.20 & 48.86 & 6.6\\ 
XGBoostLSS & 29.82 & 0.75 & $-0.33$ & 1.31 & 48.93 & 5.2\\
NGBoost & 29.39 & 0.78 & 0.48 & 1.22 & 48.93 & 6.0\\ 
cyc-GBM & 29.53 & 0.18 & $-0.50$ & 0.85 & 48.39 & 10.0\\ 
XGBoostLSSd & 29.82 & 0.75 & 0.23 & 0.76 & 48.72 & 7.4\\
\bottomrule \end{tabular}
 
\end{table}

\begin{table}[h!]
 \centering
\caption{McFadden's pseudo-$R^2$ on the test set based on the gamma deviance for the gamma models.}
\label{tab:sevgamma}
\begin{tabular}{lrrrrrr} \toprule 
\multirow{2}{*}{Model} & \multicolumn{5}{c}{Dataset} & \multirow{2}{*}{Avg. rank} \\\cmidrule{2-6}
 & WorkComp & freMTPL & BelgianMTPL & pg15training & Emcien &\\ \midrule 
 GLM & 29.41 & 0.55 & 0.45 & 5.55 & 72.49 & 6.0\\
 GAMLSS & 30.98& 0.67 & 0.71 & 2.97 & 71.96 & 8.2\\\midrule
 LightGBM & 33.24 & $-0.45$ & 0.65 & 3.90 & 68.47 & 7.0\\
XGBoost & 32.14 & $-0.04$ & 0.76 & \textbf{5.56} & 72.75 & \textbf{3.4}\\
EGBM & 32.94 & $-0.28$ & 0.53 & 4.64 & 72.65 & 6.0\\
GBM & 26.33 & $-0.09$ & 0.52 & 1.95 & 72.73 & 7.8\\
CatBoost & 31.47 & $-0.06$ & 0.81 & 5.38 & \textbf{72.85} & 3.8\\ \midrule 
XGBoostLSS & 32.72 & \textbf{0.72} & 0.61 & 5.33 & 72.71 & 4.6\\
NGBoost & \textbf{33.75} & $-0.53$ & \textbf{0.89} & 5.40 & 72.19 & 4.2\\
cyc-GBM & 25.17 & $-7.22$ & 0.79 & 4.84 & 71.97 & 6.4\\ 
XGBoostLSSd & 32.19 & $-1.30$ & 0.20 & 4.22 & 71.96 & 8.6\\
\bottomrule \end{tabular} 

\end{table}

Figures~\ref{fig:domi_freq}--\ref{fig:domi_ln}  in Appendix~\ref{app:addres} display the Murphy diagrams of the five test datasets for Poisson and gamma models. The shapes of the Murphy diagrams for BelgianMTPL and freMTPL are consistent with the ones obtained in \cite{holvoet/etal:2025}. No model is clearly dominant for BelgianMTPL in the Poisson and gamma case. In the Poisson case, we cannot clearly establish predictive dominance. Yet, NGBoost performs poorly in freMTPL and swauto. In freMTPL, cyc-GBM performs better than other algorithms, which does not reflect the result on the pseudo-$R^2$ based on the Poisson deviance. In the gamma case, no model clearly stands out, except for cyc-GBM in WorkComp and freMTPL and for GBM in pg15training. The variety of results evokes the \emph{no free lunch} theorem mentioned in \cite{hastie/etal:2009}, that is, that ``no one method dominates all others over all possible datasets''.

\subsection{Model Adequacy}
\label{sec:adequacy}

Table~\ref{tab:crpsglobal} shows the average rank in CRPS (lower is better) for the Poisson, NB2, lognormal, and gamma models. For the frequency models,  we report the uniform CRPS of the DPIT residuals. We show the values of CRPS in Tables~\ref{tab:adeqpois}--\ref{tab:crps-gamma} of Appendix~\ref{app:addres}. 

The Poisson distribution taking only a mean parameter, probabilistic and point predictions are equivalent, which leads to no clear distinction between the algorithms in terms of average rank in Table~\ref{tab:crpsglobal}. With the NB2 distribution, probabilistic prediction algorithms are, on average, less adequate than point prediction algorithms in Table~\ref{tab:crpsglobal}. This suggests that allowing the shape parameter $\phi_i$ to vary with $\mathbf{x}_i$ hinders model adequacy, at least on all the studied datasets (see Table~\ref{tab:adeqBN}). 

With the lognormal loss in Table~\ref{tab:crpsglobal}, we observe lower CRPS average ranks for probabilistic algorithms than for point algorithms. The latter aim to minimise the distance between predictions and the predicted location parameter, regardless of the scale, whereas probabilistic algorithms optimise both parameters. Such model adequacy difference does not manifest in gamma models because the gamma deviance optimised by point algorithms already entails both location and shape parameters.

\begin{table}[h]
 \centering
\caption{Average rank of CRPS for the models under the considered distributional assumptions.}
\label{tab:crpsglobal}
\begin{tabular}{lcccc} \toprule 
\multirow{2}{*}{Model} & \multicolumn{4}{c}{Distribution} \\\cmidrule{2-5}
 & Poisson & NB2 & Lognormal & Gamma \\ \midrule 
GLM & 5.0 & 5.4 & 5.6 & 6.8 \\
GAMLSS & 8.6 & 7.2  & 5.6 & 8.4 \\\midrule
LightGBM & 5.2 & \textbf{2.6} & 8.0 & 6.6\\
XGBoost & 5.2 & 2.8 & 8.2 & 6.0 \\
EGBM & 9.0 & 5.8 & 8.6 & 5.0\\
GBM & 8.8 & 5.4 & 9.8 & 9.4 \\
CatBoost & 4.8 & 3.6 & 8.2 & 3.2 \\ \midrule 
XGBoostLSS & \textbf{2.6} & 8.2 & \textbf{2.2} & 6.4 \\
NGBoost & 5.0 & 8.4 & \textbf{2.2} & \textbf{2.6}\\
cyc-GBM & 7.2 & 9.2 & 3.8 & 3.6 \\ 
XGBoostLSSd & 4.6 & 7.4 & 3.0 & 7.0\\
\bottomrule \end{tabular} 
\end{table}

\begin{figure}[t!]
\centering
    \includegraphics[width=0.95\textwidth]{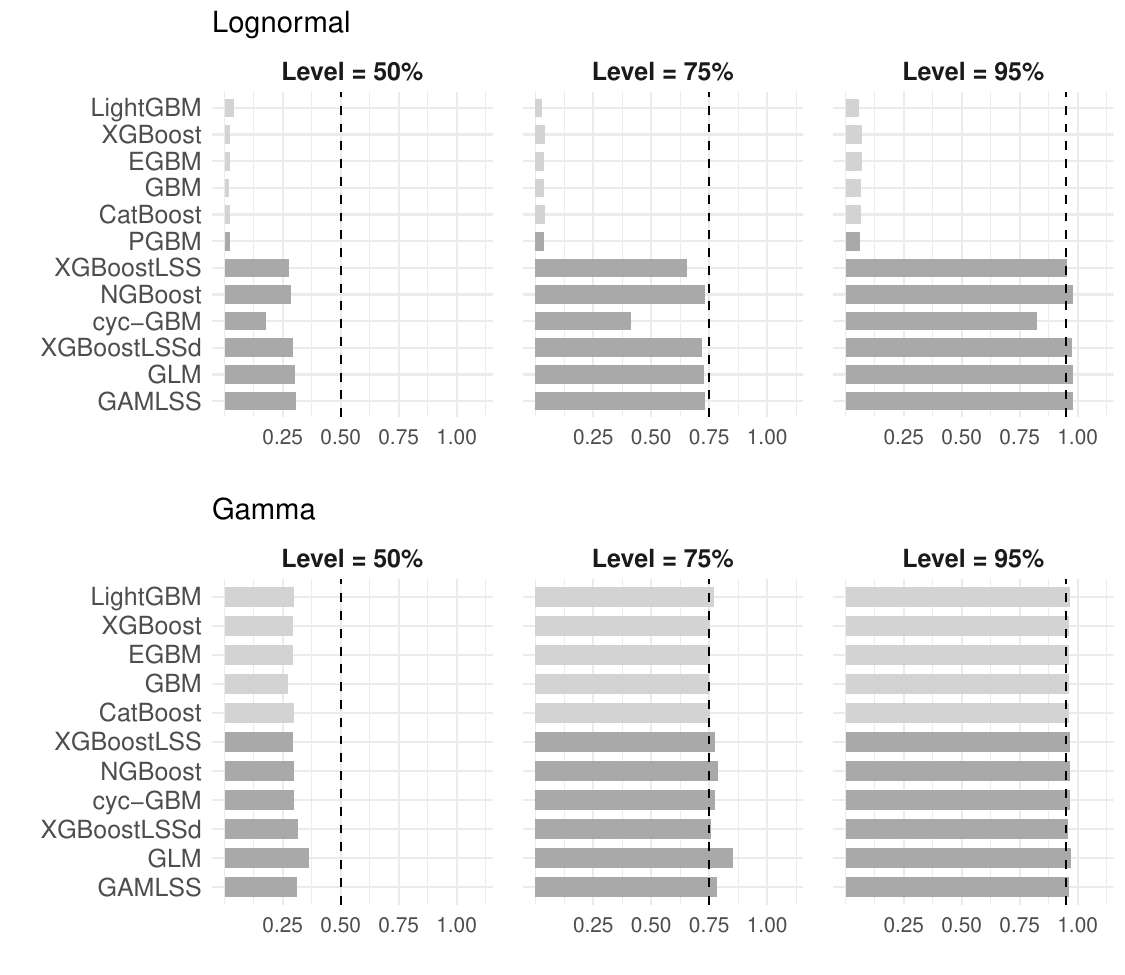}
    \caption{Coverage of CI with levels 50\%, 75\%, and 95\% on the test BelgianMTPL set for point (light) or probabilistic (dark) prediction algorithms.}
    \label{fig:cicoverage}
\end{figure}

The same dichotomy between lognormal and gamma models is diagnosed with CI coverage. Fig.~\ref{fig:cicoverage} displays CI coverage for confidence levels 50\%, 75\%, and 95\% (columns) on the BelgianMTPL dataset for point (light grey) and probabilistic (dark grey) algorithms with lognormal (top) and gamma (bottom). In each panel, a model is adequate if its bar approaches the vertical dashed line. Overall, gamma models are adequate, whereas CI predictions from lognormal point algorithms are too narrow to satisfyingly fit the BelgianMTPL severity data. A good improvement is obtained with probabilistic lognormal models, especially XGBoostLSS (with shrinkage and DART) and NGBoost. The values of CI coverage on all test datasets are in  Tables~\ref{tab:coverage-belgianmtpl}--\ref{tab:coverage-emcien} of Appendix~\ref{app:addres}. For CatBoost on the Emcien dataset, the estimated lognormal scale parameter is so large that CI coverage is 100\% at levels 75\% or 95\% (see Table~\ref{tab:coverage-lognorm}). The CI coverage of probabilistic GBDT algorithms is comparable to that of the GAMLSS benchmark.

\subsection{Portfolio balance and calibration}

Table~\ref{tab:predictedvsobservedfreq} shows the relative difference between the total rebalanced predicted and observed frequencies in Poisson models on the test set: a relative difference closer to zero is better with regards to portfolio balance. Except for freMTPL, the GLM is among the models that have the smallest relative difference with the actual number of claims. This is due to the balance property of GLMs when using the canonical link \cite[see, e.g.,][]{wuthrich/Merz:2023}. The balance property does not apply to GBDT models, so some great relative differences can be seen between the predicted and the observed frequency. We see that point prediction algorithms tend to have a smaller relative difference compared to probabilistic models. In freMTPL, BelgianMTPL, and pg15training, most of the point prediction algorithms have a positive relative difference, meaning that they predict a bigger frequency than the actual one in aggregate. For all the datasets, LightGBM exhibits the worse portfolio balance.

\begin{table}[h!]
 \centering
\caption{Relative difference (in \%) between predicted and observed frequency in Poisson models on the test set.}
\label{tab:predictedvsobservedfreq}
\begin{tabular}{lrrrrrr} \toprule 
\multirow{2}{*}{Model} & \multicolumn{5}{c}{Dataset}  & \multirow{2}{*}{Avg. rank} \\\cmidrule{2-6}
 & freMTPL & freMPL & BelgianMTPL & swauto & pg15training &\\ \midrule 
 GLM & $-0.33$ & $-0.12$ & 0.72 & -1.57 & 1.38 & 4.5\\
GAMLSS & $-0.36$ & $-0.44$ & 0.85 & -1.45 & 0.99 & 5.9\\\midrule
 LightGBM & $-53.87$ & $-49.47$ & $-11.86$ & $-19.95$ & $-19.23$ & 11.0\\
XGBoost & 0.17 & 0.02 & 0.79 & $-3.11$ & 0.59 & 4.4\\
EGBM & 0.37 & $-0.48$ & 0.80 & $-1.17$ & 1.10 & 6.0\\
GBM & 0.25 & $-1.10$ & 0.79 & $-1.79$ & 0.29 & 5.4\\
CatBoost & 0.09 & $-0.53$ & 0.75 & $-1.68$ & $-0.03$ & \textbf{3.0}\\ \midrule 
XGBoostLSS & 0.09 & $-0.66$ & 0.94 & $-1.79$ & 0.82 & 6.2\\
NGBoost & 0.19 & $-0.35$ & 0.80 & $-1.82$ & 1.56 & 6.2\\
cyc-GBM & $-0.94$ & $-1.57$ & 0.75 & $-0.83$ & 1.29 & 6.0\\ 
XGBoostLSSd & 39.63 & $2.27$ & 0.77 & $-2.00$ & 0.70 & 7.4\\
\bottomrule \end{tabular} 
\end{table}

\begin{figure}[h!]
    \centering
    \includegraphics[width=\linewidth]{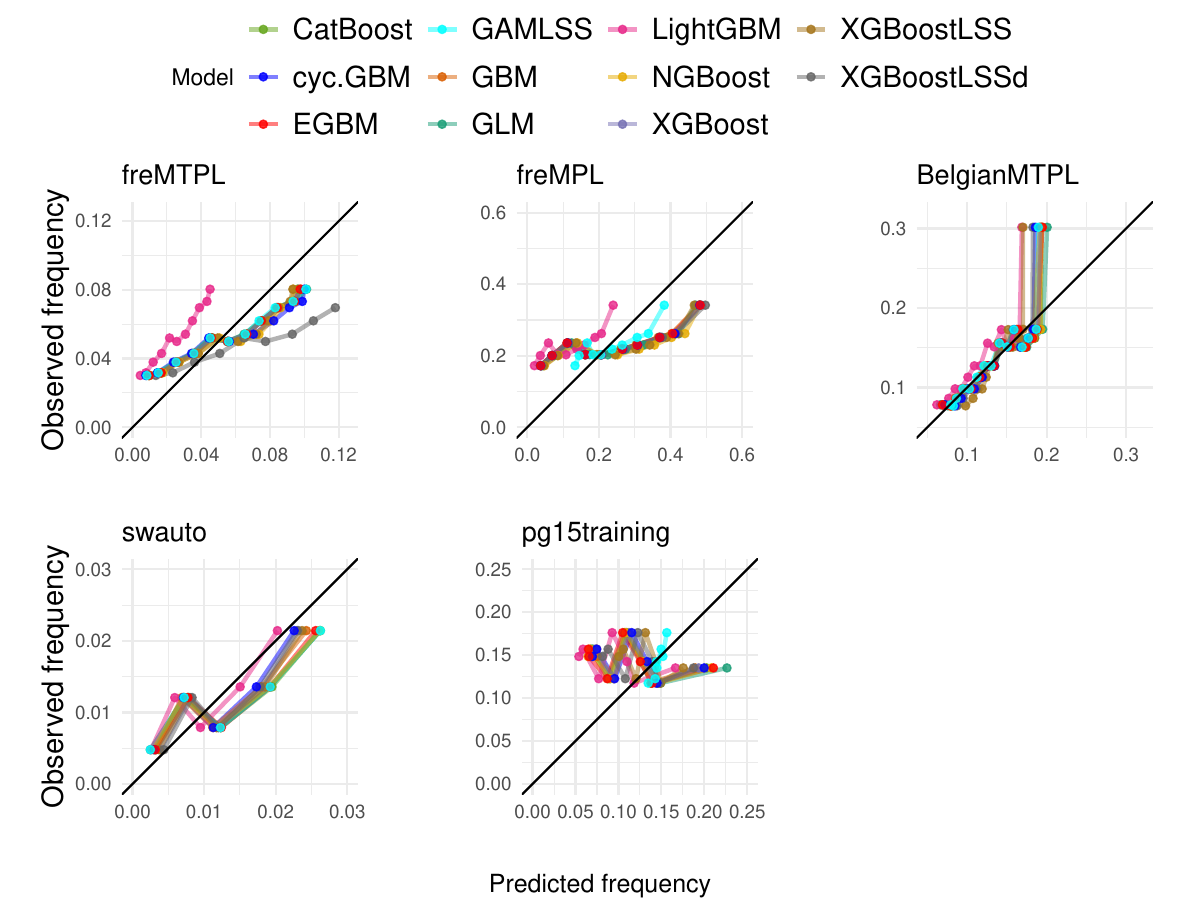}
    \caption{Calibration curves of the Poisson models on the test sets. In each bin formed from predicted frequencies, we plot the mean observed frequency against the mean predicted frequency.}
    \label{fig:calibration}
\end{figure}

Figure~\ref{fig:calibration} shows calibration curves for the rebalanced frequency models. The observations were binned according to the predicted frequency. Within each bin, we plot the mean predicted frequency against the mean observed frequency. The algorithms are best calibrated with the freMTPL and the BelgianMTPL datasets, although not ideal. There does not seem to be any major difference in calibration between the point and the probabilistic models. We see that LightGBM exhibits a slightly worse calibration than other models, especially for the French data. In Table~\ref{tab:Poissonpvalues}, we present the outcome of the auto-calibration test for the difference between concentration and Lorenz curves proposed in \cite{Denuit/etal:2024}. At a level of 1\%, the test decisions are consistent with the calibration curves in Figure~\ref{fig:calibration}, i.e., we cannot reject the null hypothesis for BelgianMTPL and swauto, whereas the models are not well calibrated for the other three datasets.

\begin{figure}[h!]
    \centering
    \includegraphics[width=\linewidth]{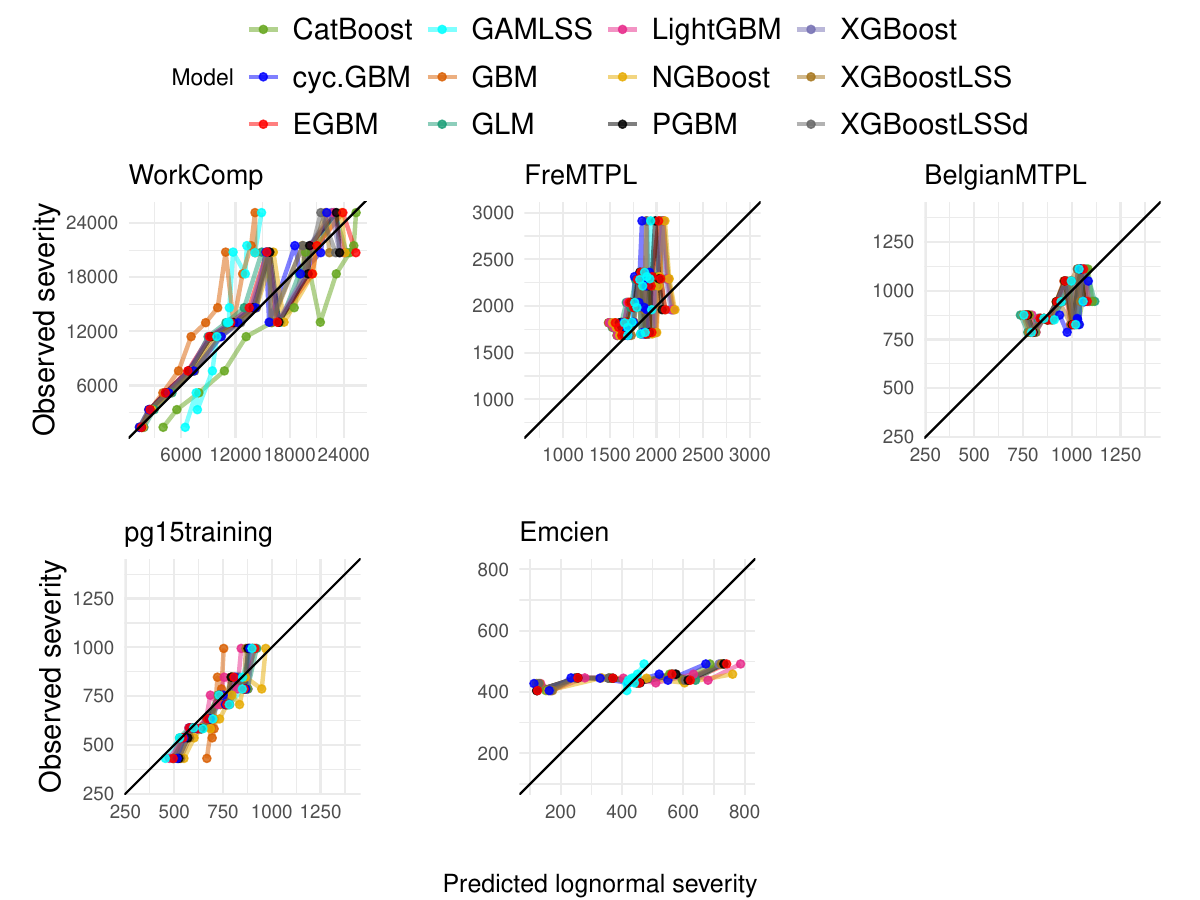}
    \caption{Calibration curves of the lognormal models on the test sets. In each bin formed from predicted severities, we plot the mean observed severity against the mean predicted severity.}
    \label{fig:calibrationln}
\end{figure}

\begin{figure}[h!]
    \centering
    \includegraphics[width=\linewidth]{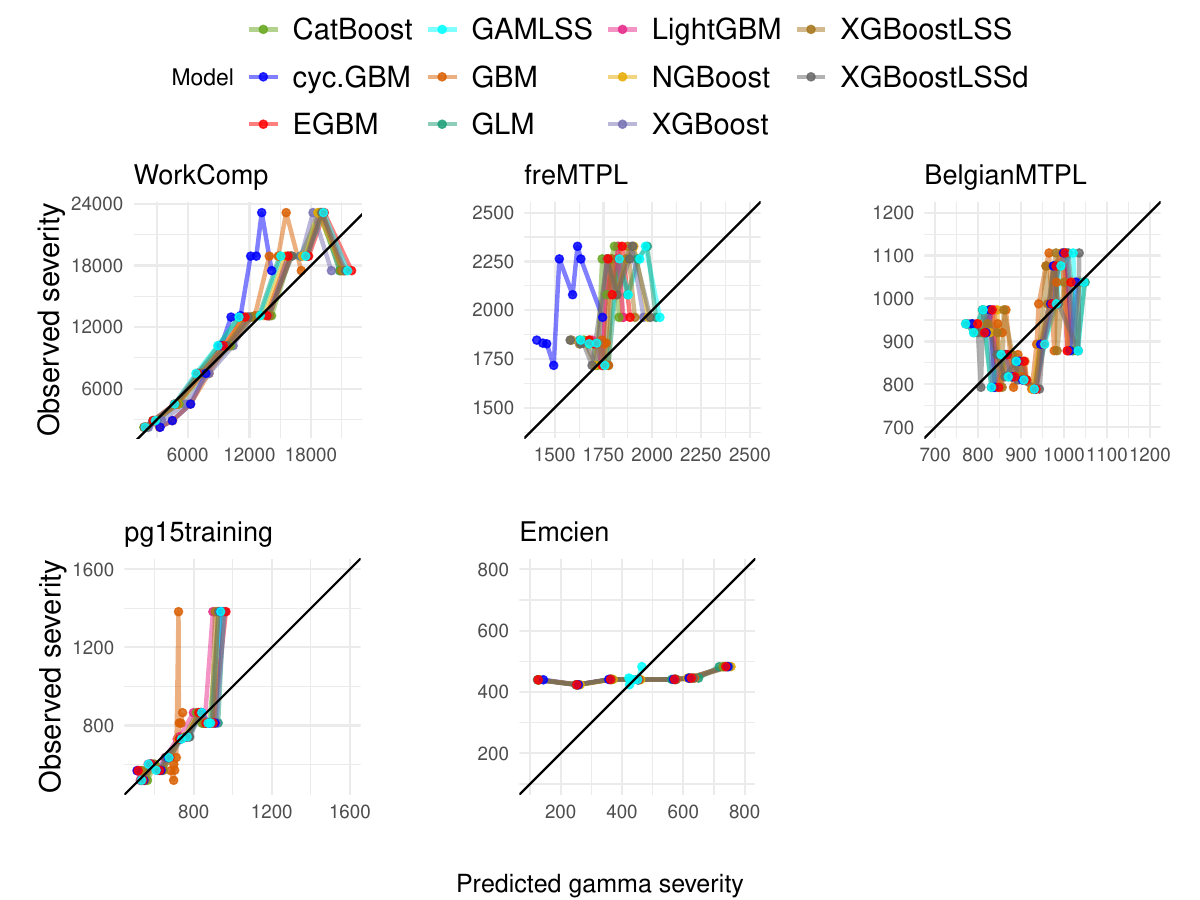}
    \caption{Calibration curves of the gamma models on the test sets. In each bin formed from predicted severities, we plot the mean observed severity against the mean predicted severity.}
    \label{fig:calibrationgamma}
\end{figure}

Figures~\ref{fig:calibrationln} and~\ref{fig:calibrationgamma} show the calibration curves for rebalanced severity models. We see that these models are, in some cases, poorly calibrated. In the lognormal case, freMTPL, BelgianMTPL, and pg15training display a calibration that is fairly aligned with the diagonal. In the WorkComp dataset, GBM and CatBoost are respectively an underestimation and an overestimation of the average severity per bin. In the gamma case, pg15training models are reasonably well calibrated. We see that cyc-GBM is poorly calibrated for WorkComp and freMTPL. For both distributions, the severity models for the Emcien dataset are poorly calibrated. Tables~\ref{tab:LNpvalues} and~\ref{tab:gammapvalues} display the $p$-values of the auto-calibration test, and give a more precise outlook on calibration.

\begin{table}[h!]
\centering
\caption{$p$-values of the auto-calibration test for the Poisson models.}
\label{tab:Poissonpvalues}
\begin{tabular}{lrrrrr} \toprule 
\multirow{2}{*}{Model} & \multicolumn{5}{c}{Dataset} \\\cmidrule{2-6}
& freMTPL & freMPL & BelgianMTPL & swauto & pg15training \\ \midrule 
GLM & 0&  0& 0.493 &   0.087& 0\\
GAMLSS & 0 & 0 & 0.261 & 0.086  & 0.721 \\\midrule
LightGBM & 0&  0& 0 &  0.024&  0\\
XGBoost & 0&  0&  0.798&  0.181&  0\\
EGBM & 0&  0&  0.328&  0.248&  0\\
GBM & 0&  0&  0&  0.244&  0\\
CatBoost &  0& 0& 0.840 & 0.262&  0\\ \midrule 
XGBoostLSS &  0&  0& 0.022 &  0.198&  0\\
NGBoost &  0&  0& 0.300 & 0.130& 0\\
cyc-GBM & 0&  0& 0.775 &  0.017& 0\\ 
XGBoostLSSd &  0&  0& 0.535 &  0.403&  0\\
\bottomrule \end{tabular} 
\end{table}

\begin{table}[h!]
\centering
\caption{$p$-values of the auto-calibration test for the lognormal models.}
\label{tab:LNpvalues}
\begin{tabular}{lrrrrr} \toprule 
\multirow{2}{*}{Model} & \multicolumn{5}{c}{Dataset} \\\cmidrule{2-6}
& WorkComp & freMTPL & BelgianMTPL & pg15training & Emcien \\ \midrule 
GLM & 0.342 & 0.008 &  0.142&  0.570 & 0\\
GAMLSS & 0.001 & 0.021 &  0.107&  0.060 & 0.130 \\\midrule
LightGBM & 0.013 & 0.014 & 0.352 & 0.037 & 0 \\
XGBoost & 0.039 & 0.018 & 0.176 & 0.247 & 0 \\
EGBM & 0.010 & 0.036 & 0.360 & 0.635 & 0 \\
GBM & 0 & 0.041 & 0.158 & 0.112 & 0 \\
CatBoost & 0 & 0.035 & 0.511 & 0.815 & 0 \\ \midrule 
XGBoostLSS & 0.023 & 0.065 & 0.001 & 0.360 & 0 \\
NGBoost & 0.147 & 0.043 & 0.281 & 0 & 0 \\
cyc-GBM & 0& 0 & 0 & 0.337 & 0\\ 
XGBoostLSSd & 0.005 & 0.031 & 0.014 & 0.060 & 0 \\
\bottomrule \end{tabular} 
\end{table}

\begin{table}[h!]
\centering
\caption{$p$-values of the auto-calibration test for the gamma models.}
\label{tab:gammapvalues}
\begin{tabular}{lrrrrr} \toprule 
\multirow{2}{*}{Model} & \multicolumn{5}{c}{Dataset} \\\cmidrule{2-6}
& WorkComp & freMTPL & BelgianMTPL & pg15training & Emcien \\ \midrule 
GLM & 0.343 & 0.047 &  0.579&  0.537 & 0\\
GAMLSS & 0.222 & 0.017 & 0.734 & 0.252  & 0.103 \\\midrule
LightGBM & 0.794 & 0.006 & 0.852 & 0.134 & 0 \\
XGBoost & 0.332 & 0.003 & 0.857 & 0.350 & 0 \\
EGBM & 0.179 & 0.018 & 0.797 & 0.373 & 0 \\
GBM & 0.034 & 0.045 & 0.379 & 0.039 & 0 \\
CatBoost & 0.897 & 0.003 & 0.658 & 0.493 & 0 \\ \midrule 
XGBoostLSS & 0.596 & 0.018 & 0.720 & 0.570 & 0 \\
NGBoost & 0.190 & 0.002 & 0.732 & 0.554 & 0 \\
cyc-GBM & 0& 0 & 0.840 & 0.604 & 0\\ 
XGBoostLSSd & 0.310 & 0.001 & 0.328 & 0.221 & 0 \\
\bottomrule \end{tabular} 
\end{table}

\newpage
\subsection{Comparison of the resulting tariff structures}
Tables~\ref{tab:giniindex_poissongamma} and~\ref{tab:giniindex_poissongammafrench} compare the resulting tariff structures through the Gini index on the BelgianMTPL and freMTPL test sets, respectively. Computed with the ordered Lorenz curves introduced by \cite{frees/etal:2014}, the Gini index is twice the area between this curve and the 45 degree line or, as explained by \cite{frees/etal:2014}, twice the average profit for an insurance company over all the possible strategies with respect to portfolio composition. A high Gini index implies that the model performs well in segmenting good and bad risks. For each row of Tables~\ref{tab:giniindex_poissongamma} and~\ref{tab:giniindex_poissongammafrench}, the column with the highest Gini index indicates the model that improves the most the row model. As outlined in \cite{frees/etal:2014}, the preferred model is the one that is the least improved. Hence, a min-max approach is taken by first highlighting the row-wise highest Gini indices and then selecting the row with the smallest highlighted value. We study the tariff structures resulting from the combination of the Poisson and gamma rebalanced predictions. With respect to the Gini index, we find that GBM performed best for the Belgian data and that CatBoost is the best for freMTPL. Interestingly, the latter dataset contains a high cardinality categorical variable.

\section{Discussion}
\label{sec:discussion}

In this study, we gather the recent developments of GBDT for point and probabilistic predictions in a unified notation. We adopt an actuarial standpoint as GBMs are considered as an alternative to GLMs in general insurance ratemaking and reserving. GBDT approaches compete on five actuarial datasets for claim frequency and severity modelling, based on computational efficiency, predictive performance and model fit.  

There is no one-size-fits-all solution, but some algorithms stand out. LightGBM can more efficiently output predictions that are as precise as those of other algorithms. Also, CatBoost can improve predictive performance in the presence of high cardinality categorical variables, which are frequent in insurance. Finally, model adequacy may be enhanced by using probabilistic GBDT without hurting much predictive performance. We find that XGBoostLSS is the most computationally efficient probabilistic GBDT algorithm and has competitive performance. 

In actuarial applications, model interpretability matters. Unlike GLMs, GBDT generally does not lead to interpretable models. Tools such as variable importance or a partial dependence plot (PDP) can help in explaining GBDT model outputs, as illustrated with the BelgianMTPL dataset in \cite{Henckaerts/etal:2021}. Modelling multiple distributional parameters based on covariates adds an additional layer of complexity, as these tools need to be applied on each of them. Moreover, warnings about PDPs are issued in~\cite{xin/etal:2024-pdp}: with adversarial attacks, they can be ``deceptive''. 

\begin{figure}[t!] 
    \centering
    \includegraphics[width=0.8\linewidth]{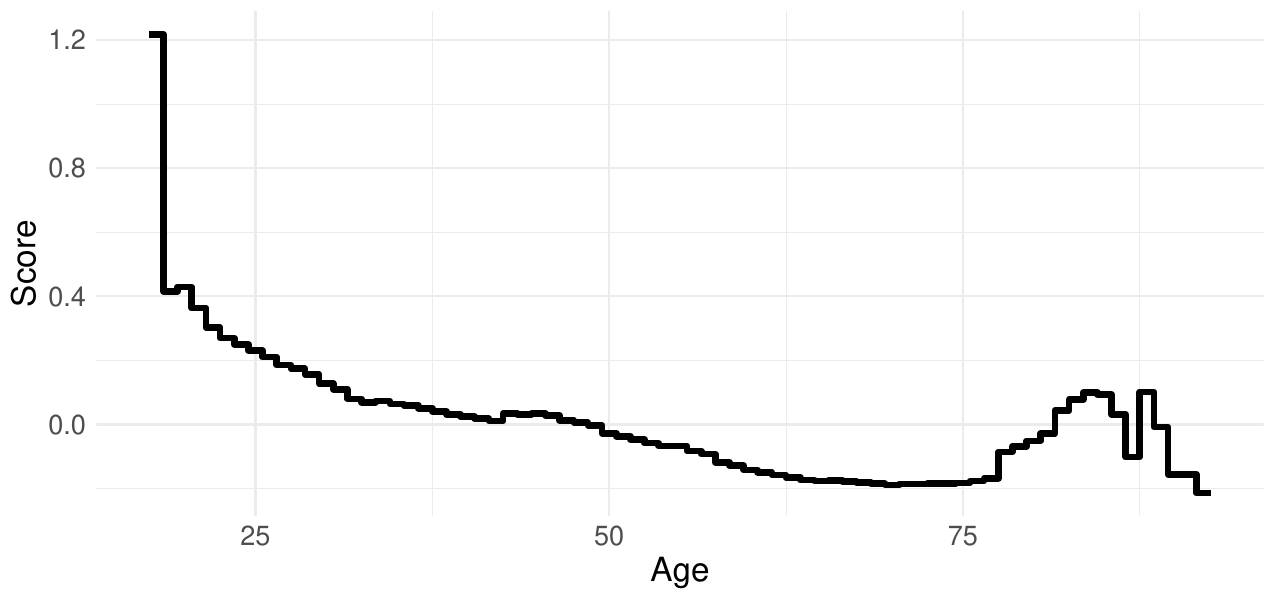}
    \caption{Main effect for the variable \texttt{ageph} in the Poisson EGBM on BelgianMTPL.}
    \label{fig:explication_ageph}
\end{figure}

The fully interpretable algorithm EGBM achieves competitive predictive performance compared to black box GBDT algorithms. It outputs regression parameters in a lookup table, allowing to retrieve feature and interaction effects. Fig.~\ref{fig:explication_ageph} shows the main effect $f_{\texttt{ageph}}^M$ in the BelgianMTPL Poisson EGBM: its shape mimics the corresponding smooth GAMLSS effect in \cite{klein/etal:2014}. Interestingly, post-hoc adjustments can be made, allowing actuaries to manually handle the EGBM model's extrapolation behaviour in rarely observed segments, e.g., when \texttt{ageph} is smaller than 20 or larger than 75 in Fig.~\ref{fig:explication_ageph}. This is impossible with other opaque GBDT methods, as global or local explanations cannot be used to adjust predictions post-hoc. 

\begin{figure}[t!] 
    \centering
    \includegraphics[width=0.8\linewidth]{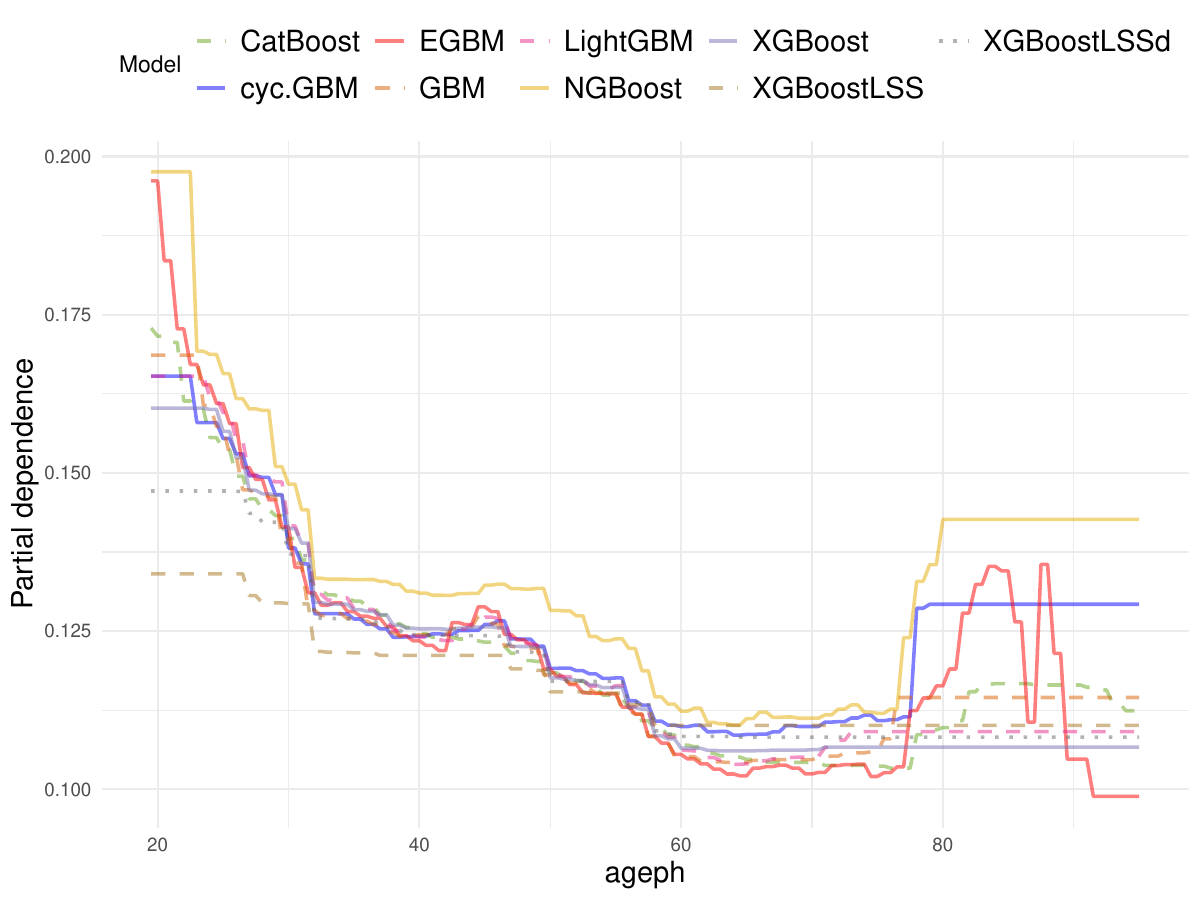}
    \caption{Partial dependence plots for the variable \texttt{ageph} in the Poisson EGBM on BelgianMTPL.}
    \label{fig:pdp_ageph}
\end{figure}

Interpretation tools such as individual conditional expectation (ICE) plots \citep{Goldstein/etal:2015} or partial dependence plots (PDPs) \cite{Friedman:2001} are known, and widely used, to gain insights on the effects included in black-box models. We can compare the insights from these methods, which are approximate, to the exact insights, which are readily available in the output EGBM model. Figure~\ref{fig:pdp_ageph} shows the PDPs of the \texttt{ageph} variable for all the Poisson models on BelgianMTPL. Except for NGBoost and cyc-GBM, we see that the partial dependence has a similar shape to the one for the EGBM model. For $\texttt{ageph} > 60,$ the XGBoostLSS and XGBoostLSSd models do not exhibit the same increasing partial dependence as the other models. The shape of the EGBM partial dependence curve follows directly from the main effect displayed on Figure~\ref{fig:explication_ageph}.

\begin{figure}[h!] 
    \centering
    \includegraphics[width=\linewidth]{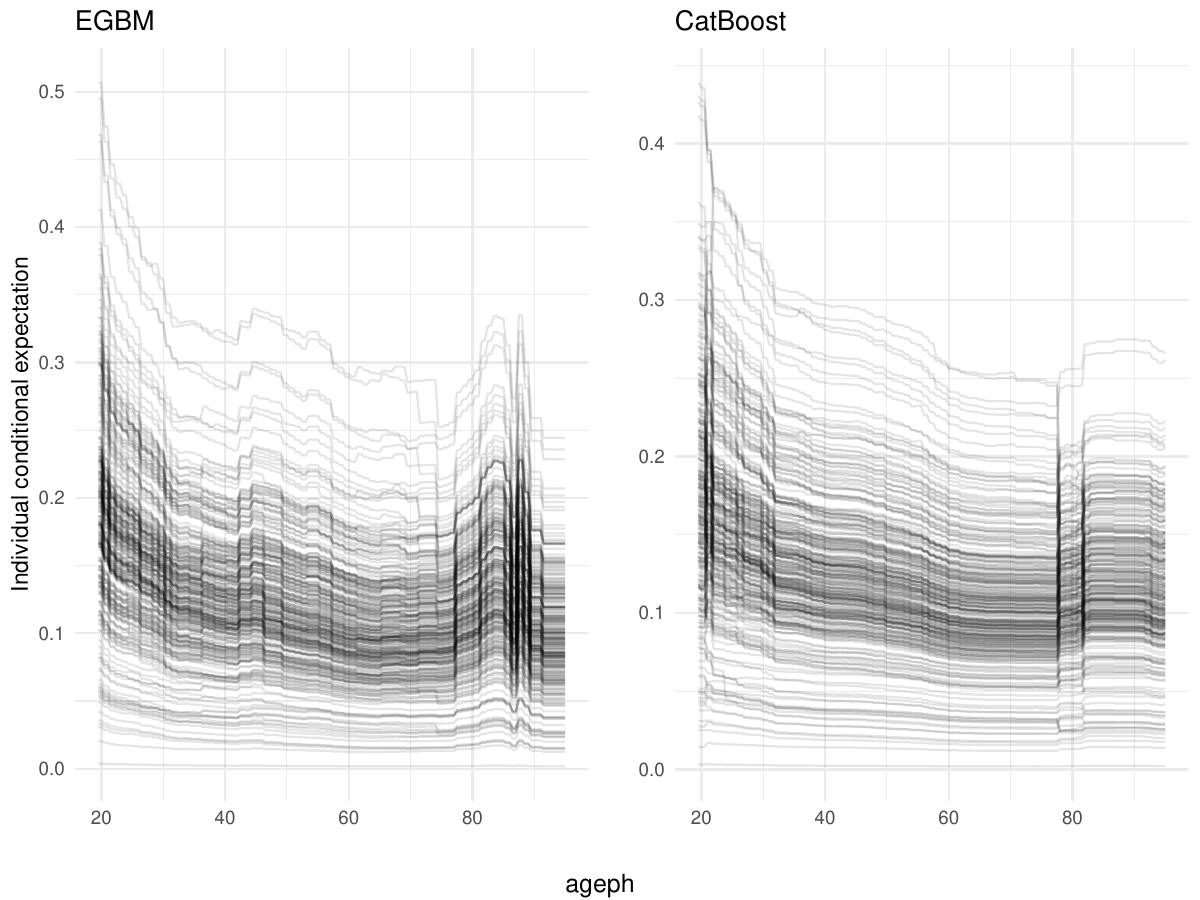}
    \caption{Individual conditional expectation plots for the EGBM and CatBoost Poisson models on BelgianMTPL based on the \texttt{ageph} variable. A subsample of 300 observations is taken to enhance readability.}
    \label{fig:ice_ageph}
\end{figure}

PDPs are an average over all the observations of the ICE plots. Interactions between the variable of interest (\texttt{ageph} in this example) and the other ones can create distortions in the PDPs, so they need to be analysed with care. Xin et al. (2024) \cite{xin/etal:2024-pdp} provide examples of how PDPs can be misleading. Figure~\ref{fig:ice_ageph} shows the ICE plots of the \texttt{ageph} variable for the EGBM and CatBoost BelgianMTPL Poisson models. These plots are based on a random subsample of 300 observations to enhance readability. The ICE plots for EGBM seem to be more refined than the ones for the CatBoost model. In both cases, we see ICE curves that intersect, meaning that there are interactions between \texttt{ageph} and other variables. For the CatBoost model, Friedman's $H$-statistic could be used to find the significant interactions in the model \citep[see][]{friedman:2008}. For EGBM, we know directly from the model output that, above from the \texttt{ageph} main effect, interactions \texttt{ageph \& sex}, \texttt{ageph \& bm}, \texttt{ageph \& power}, and \texttt{ageph \& agec} were learned on the residuals, as selected 
 by the FAST algorithm introduced in \cite{lou:2013:GA2M}. It is also possible to extract the exact interaction effects from the EGBM model output to gain a complete understanding of the effect of \texttt{ageph} in claim frequency prediction for the BelgianMTPL data. The same insights cannot be obtained exactly from any other GBDT black-box model.

Even if multi-parametric GBDT methods allow to predict a distinct value of distributional parameter per observation, it may be appropriate to keep some parameters constant. For example, an individual's characteristics may change the behaviour of its risk, but not as dramatically as the possible variations in the shape parameter of a distribution. Under the gamma distributional assumption, we find that using a global shape parameter does not hinder much model adequacy. The shape of the gamma density is dramatically different when the shape parameter crosses one, which could indeed warrant a single shape parameter.

In NB2 models, we find that a global dispersion parameter enhances model adequacy compared to a covariate-specific dispersion parameter. One could argue that the overdispersion of a negative binomial should not vary per observation. Conceptually, overdispersion must be diagnosed with more than one observation, which could warrant to model overdispersion collectively, or over a small number of risk classes. Keeping a distributional parameter constant can be done post-hoc with point predictions (as described in Section~\ref{sec:frompointtoprob}) or iteratively with parameter-specific hyperparameters \citep[as in cyc-GBM][]{Delong/etal:2023}: setting tree depth to zero yields a global parameter estimation.

The observations we make on portfolio balance and auto-calibration emphasise that boosting algorithms do not have the desirable balance property of GLMs \citep[see, e.g.,][]{wuthrich:2020} even if they can output probabilistic predictions. To correct this problem, an auto-calibration procedure for insurance pricing is proposed in \cite{Denuit/etal:2021}, and its performance with GBM is demonstrated. This procedure, along with one of the studied GBDT algorithms, could improve the adequacy of predictions for general insurance ratemaking while keeping the balance property, as $\hat{\mu}(\mathbf{X})=E\{Y|\hat{\mu}(\mathbf{X})\}$ is ensured by auto-calibration. Auto-calibration may also be improved by applying isotonic recalibration on the chosen GBDT model without hindering its segmentation performance as proposed in \cite{wuthrich/ziegel:2024}.

\section*{Acknowledgments}

We are thankful to an anonymous insurance company for useful discussions. We thank two anonymous reviewers whose comments helped improve this comparative analysis. We acknowledge the help of Lu Yang for the implementation of DPIT residuals for Poisson models and Olivier Côté for his comments. This research was funded by the Natural Sciences and Engineering Research Council of Canada (CRDPJ 515901-17, RGPIN-2019-04190) and the Chaire d'actuariat of Université Laval.

\appendix

\section{Tuning strategy}
\label{app:tuning}
We follow Algorithm~\ref{alg:scheme} to tune hyperparameters with the five datasets used in claim frequency and severity modelling. We emphasise that all performance metrics are computed on a test set $\mathcal{D}_{\mathrm{test}}$ which is never seen in the tuning and the training process. To be specific, for dataset $k$, metric $\psi$ and resulting model output from Algorithm~\ref{alg:scheme}, we compute
$$
\frac{1}{|\mathcal{D}_{\mathrm{test},k|}}\sum_{i \in \mathcal{D}_{\mathrm{test},k}} \psi\{y_i, f_{k, \theta_k^*}(\mathbf{x}_i)\}.
$$
Data partition is identical for all the algorithms.

Grid search is used for the maximum number of boosting iterations $M$ and the tree depth $d.$ We search over the same grid for these two hyperparameters across algorithms, but this grid is augmented by specific hyperparameters when necessary. We set the (constant) learning rate at 0.01, the sampling proportion at $\delta = 0.75$, and the minimum number of samples per leaf at 1\% of the number of training observations.

\begin{algorithm}[t]
    \SetKwFunction{isOddNumber}{isOddNumber}
    \SetKwInOut{KwIn}{Input}
    \SetKwInOut{KwOut}{Output}

    \KwIn{Datasets $\mathcal{D}_1, \mathcal{D}_2, \mathcal{D}_3, \mathcal{D}_4,$ and $\mathcal{D}_5,$ tuning grid \texttt{grid}, algorithm \texttt{algo}, and loss function $\mathcal{L}$.}
    \For{$k = 1$ \KwTo $5$}{
    Sample $\mathcal{D}_{\mathrm{train}, k},$ a random subset of 85\% of observations in $\mathcal{D}_k$.
    
    Assign to $\mathcal{D}_{\mathrm{test}, k}$ the remaining 15\% of observations in $\mathcal{D}_k$.
    
    Sample $\mathcal{D}_{\mathrm{hyp\_train}, k},$ a random subset of 80\% of observations in $\mathcal{D}_{\mathrm{train}, k}.$

    Assign to $\mathcal{D}_{\mathrm{val}, k}$ the remaining 20\% of observations in $\mathcal{D}_{\mathrm{train}, k}$.
    \BlankLine
    \For{parameter combination $\theta$ in \texttt{grid}}{
   Train model $f_{k, \theta}$ using \texttt{algo} with $\theta$ on $\mathcal{D}_{\mathrm{hyp\_train}, k}.$
   
    Compute \texttt{score}$_{k, \theta} = \frac{1}{|\mathcal{D}_{\mathrm{val}, k}|} \sum_{i\in \mathcal{D}_{\mathrm{val}, k}} \mathcal{L}\{y_i, f_{k, \theta}(\mathbf{x}_i)\}.$ 
    }
    \BlankLine
       Choose the combination $\theta_k^*$ which returns minimal $\texttt{score}_{k, \theta}$. 
    \BlankLine
   Train model $f_{k, \theta^*}$ using \texttt{algo} with $\theta_k^*$ on $\mathcal{D}_{\mathrm{train}, k} = \mathcal{D}_{\mathrm{hyp\_train}, k} \cup \mathcal{D}_{\mathrm{val}, k}.$  }
   \textbf{Return : } For $k \in \{1, \ldots, 5\},$ optimal hyperparameters $\theta_k^*$, trained model $f_{k, \theta^*_k}$.
   \caption{Grid search scheme for tuning}
    \label{alg:scheme}
\end{algorithm}

\section{Additional results}
\label{app:addres}

In Table~\ref{tab:sevefficiency}, we show the training time in seconds for lognormal models. The average ranks are consistent with what we observe in Poisson models. 

\begin{figure}[h!]
    \centering
    \includegraphics[width=\linewidth]{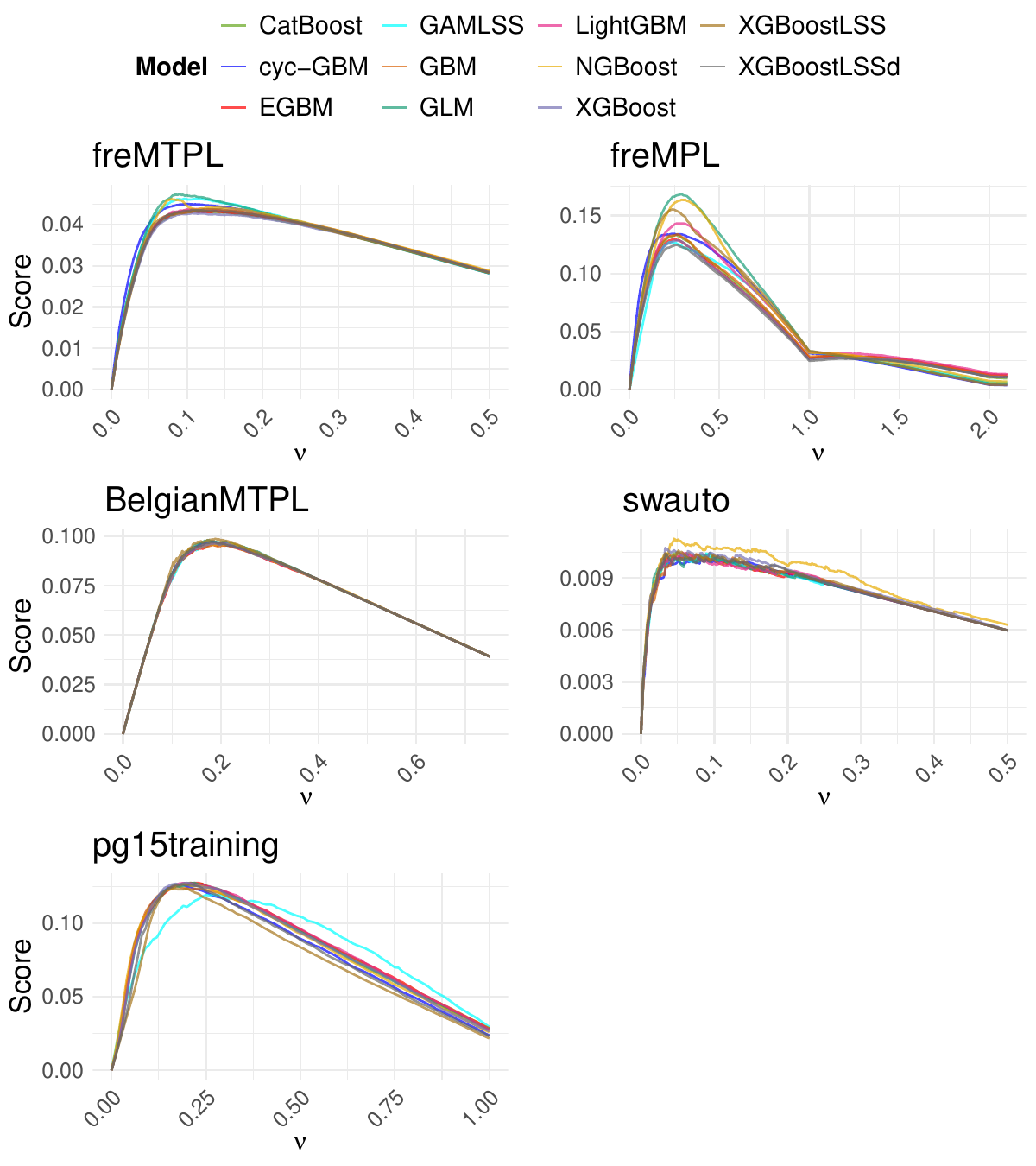}
    \caption{Murphy diagrams comparing the predictive dominance of the Poisson models on the test sets for various values of $\nu$ according to the elementary scoring function $S_{\nu}.$}
    \label{fig:domi_freq}
\end{figure}

\begin{figure}
    \centering
    \includegraphics[width=\linewidth]{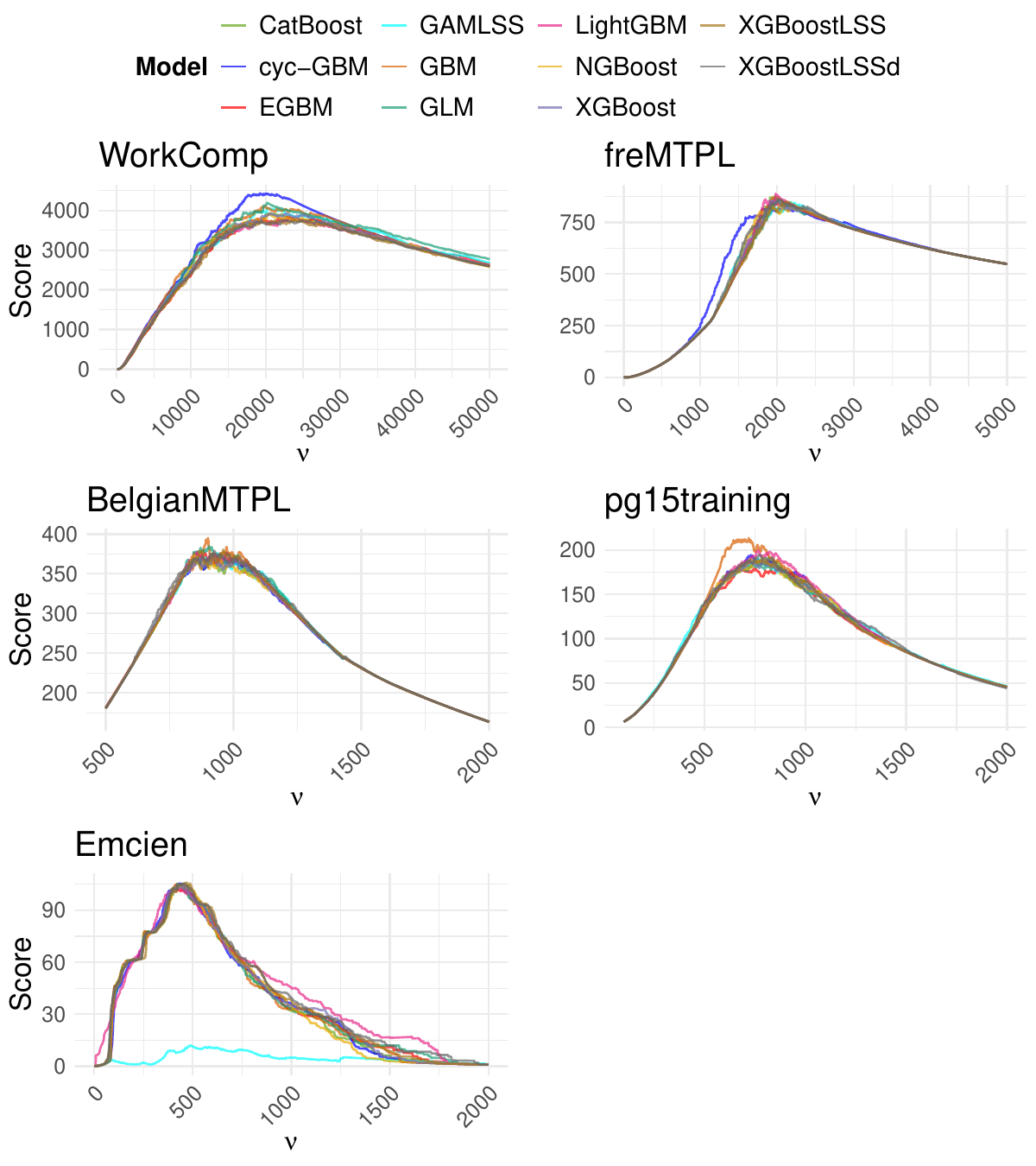}
    \caption{Murphy diagrams comparing the predictive dominance of the gamma models on the test sets for various values of $\nu$ according to the elementary scoring function $S_{\nu}.$}
    \label{fig:domi_sev}
\end{figure}

\begin{figure}
    \centering
    \includegraphics[width=\linewidth]{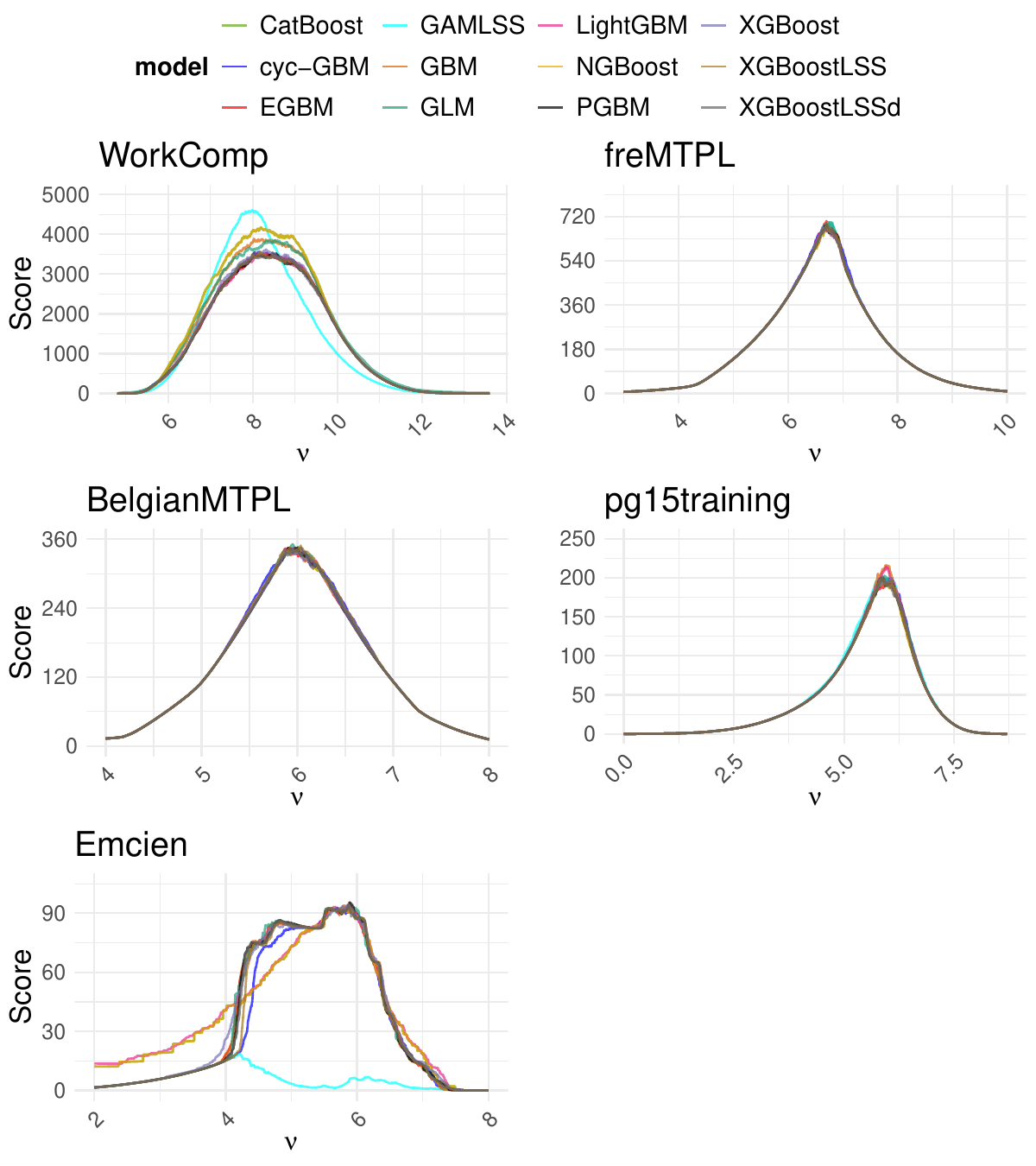}
    \caption{Murphy diagrams comparing the predictive dominance of the lognormal models on the test sets for various values of $\nu$ according to the elementary scoring function $S_{\nu}.$}
    \label{fig:domi_ln}
\end{figure}

Figures~\ref{fig:domi_freq}---\ref{fig:domi_ln} present the Murphy diagrams for the Poisson, gamma, and lognormal models. We note that the GAMLSS clearly dominates all other models on the Emcien dataset, which means that it may be close to the data generating mechanism. The behaviour of the Emcien dataset with respect to the coverage of confidence intervals in Table~\ref{tab:coverage-emcien} resembles that of the simulated WorkComp dataset. However, there is no indication in \cite{emcien} on whether Emcien is synthetic or real.

The adequacy of claim frequency models is assessed through DPIT residuals. These residuals should follow a uniform distribution if the model is adequate \citep{yang:2024}. Tables~\ref{tab:adeqpois}--\ref{tab:adeqBN} display the CRPS values for the Poisson and the NB2 models, respectively.

\begin{table}[h!]
\centering
\caption{Training time in seconds for lognormal models.}
\label{tab:sevefficiency}
\begin{tabular}{lrrrrrr}
\toprule
\multirow{2}{*}{Model} & \multicolumn{5}{c}{Dataset} & \multirow{2}{*}{Avg. rank} \\\cmidrule{2-6}
& WorkComp & freMTPL & BelgianMTPL & pg15training & Emcien & \\ \midrule
    LightGBM  & 7 & 2 & 3 & 2 & 2 & \textbf{1.0}\\
    XGBoost & 21 & 5 & 15 & 18 & 4 & 2.2\\
    EGBM  & 26 & 6 & 24 & 78 & 6 & 3.8\\ 
    GBM  & 78 & 40 & 19 & 25 & 5 & 4.0\\
    CatBoost  & 43 & 16 & 28 & 83 & 29 & 4.0\\ \midrule
    PGBM & 22 & 5 & 16 & 20 & 7 & 1.2\\
    XGBoostLSS  & 46 & 16 & 36 & 41 & 9 & 2.2\\
    NGBoost  & 495 & 38 & 46 & 121 & 66 & 2.6\\
    cyc-GBM  & 743 & 240 & 266 & 1~188 & 105 & 4.6\\
    XGBoostLSSd  & 556 & 278 & 251 & 1~082 & 399 & 4.4\\
\bottomrule
\end{tabular}
\end{table}

    \begin{table}[h!]
        \centering
            \caption{Uniform CRPS ($\times 10^2$) of the DPIT residuals in Poisson models on the test set.}
  \begin{tabular}{lcccccc}
    \toprule
\multirow{2}{*}{Model} & \multicolumn{5}{c}{Dataset} & Avg. rank\\\cmidrule{2-6}
    & freMTPL &freMPL & BelgianMTPL  & swauto & pg15training\\ \midrule
    GLM & 16.173 & \textbf{15.89} &  16.495 & 16.669 &  16.67 & 5.0\\
    GAMLSS & 16.301  & 16.61 & 16.499 & 16.674& 16.79 & 8.6\\ \midrule
    LightGBM & 16.299 & 16.33& 16.487& 16.666& 16.64 & 5.2\\
    XGBoost  & 16.302 & 16.03& 16.286& 16.671& 16.43 & 5.2\\ 
    EGBM     & 16.354 & 16.62& 16.494& 16.683& 16.73 & 9.0\\
    GBM      & 16.332 & 16.22 & 16.502& 16.685& 16.80 & 8.8\\
    CatBoost & 16.294 & 16.32& 16.496& 16.665& 16.62 & 4.8\\
    \midrule
    XGBoostLSS & 16.298 & 15.96 & \textbf{16.283} & 16.672& \textbf{16.38} & \textbf{2.6}\\
    NGBoost & \textbf{16.098} & 15.97 & 16.461 & 16.691& 16.70 & 5.0\\
    cyc-GBM & 16.628 & 16.30 & 16.610& 16.665& 16.65 & 7.2\\
    XGBoostLSSd & 16.323 & 16.60& 16.452& \textbf{16.649}& 16.40 & 4.6\\
    \bottomrule
    \end{tabular}
    \label{tab:adeqpois}
    \end{table}

    \begin{table}[h!]
    \centering
    \caption{Uniform CRPS ($\times 10^2$) of the DPIT residuals in NB2 models on the test set.}
  \begin{tabular}{lcccccc}
    \toprule
\multirow{2}{*}{Model} & \multicolumn{5}{c}{Dataset} & Avg. rank\\\cmidrule{2-6}
    & freMTPL &freMPL & BelgianMTPL  & swauto & pg15training\\ \midrule
    GLM & 16.373 & 16.26 & 16.486 & 16.6702 & 17.789 & 5.4\\
    GAMLSS & 16.883 & 16.32 & 16.472 & 16.7543 & 16.673 & 7.2\\ \midrule
    LightGBM & 16.316 & 15.86 & 16.491 & 16.6780& \textbf{16.577} & \textbf{2.6}\\
    XGBoost  & \textbf{16.320} & 16.01 & \textbf{16.488} & \textbf{16.5753}& 16.599 & 2.8\\
    EGBM      & 16.360 & 16.19 & 16.496& 16.6910& 16.681 & 5.8\\
    GBM      & 16.345 & \textbf{15.81} & 16.505& 16.6856& 16.732 & 5.4\\
    CatBoost & 16.321 & 15.91 & 16.500 & 16.6729& 16.580 & 3.6\\
    \midrule
    XGBoostLSS & 16.734 & 17.38 & 16.668& 16.6989& 16.636 & 8.2\\
    NGBoost & 16.710 & 17.30 & 16.672& 16.6988& 16.682 & 8.4\\
    cyc-GBM & 16.898 & 17.05 & 16.779 & 16.7396& 16.669 & 9.2\\
    XGBoostLSSd & 16.745 & 17.36 & 16.609 & 16.7214& 16.646 & 7.4\\
    \bottomrule
    \end{tabular}
    \label{tab:adeqBN}
    \end{table}

\begin{table}[h!]
\centering
     \caption{CRPS in lognormal models on the test set.}
\begin{tabular}{lcccccc}
\toprule
\multirow{2}{*}{Model} & \multicolumn{5}{c}{Dataset} & Avg. ranl\\\cmidrule{2-6}
     & WorkComp & freMTPL & BelgianMTPL & pg15training & Emcien \\ \midrule
     GLM & 0.625 & 0.665 & 0.807 & 0.674 & 0.221 & 5.6\\
    GAMLSS & 0.789 & 0.662 & 0.806& 0.689 & 0.195 & 5.6\\ \midrule
    LightGBM  & 0.586 & 0.768 & 1.157 & 0.810 & 0.216 & 8.0\\
    XGBoost & 0.592 & 0.779 & 1.133 & 0.816 & 0.263 & 8.2\\
    EGBM  & 0.588 & 0.791 & 1.140 & 0.818 & 0.264 & 8.6\\
    GBM  & 0.619 & 0.785 & 1.142 & 0.868 & 0.262 & 9.8\\
    CatBoost  & 0.652 & 0.790 & 1.140 & 0.808 & 0.210 & 8.2\\\midrule
    PGBM  & 0.746 & 0.803 & 1.146 & 0.865 & 0.238 & 6.4\\
    XGBoostLSS  & \textbf{0.573} & 0.659 & 0.816 & \textbf{0.672} & \textbf{0.187} & \textbf{2.2}\\
    NGBoost  & 0.580 & \textbf{0.658} & \textbf{0.805} & 0.673 & 0.189 & \textbf{2.2}\\
    cyc-GBM  & 0.578 & 0.663 & 0.860 & 0.675 & 0.188 & 3.8\\
     XGBoostLSSd  & 0.574 & 0.659 & 0.806 & 0.679 & 0.190 & 3.0\\
\bottomrule
\end{tabular}
\label{tab:crps-lognorm}
\end{table}
For severity models, the predicted distribution can be used directly to assess model adequacy with the CRPS. Tables~\ref{tab:crps-lognorm}--\ref{tab:crps-gamma} display the CRPS for the severity models. 
\begin{table}[h!]
\centering
     \caption{CRPS in gamma models on the test set.}
\begin{tabular}{lcccccc}
\toprule
\multirow{2}{*}{Model} & \multicolumn{5}{c}{Dataset} & Avg. rank\\\cmidrule{2-6}
& WorkComp & freMTPL & BelgianMTPL & pg15training & Emcien\\ \midrule
    GLM & 7~023 &1~343.4  & 544.8 & 335.91 & 63.3 & 6.8\\
    GAMLSS  &7~045 & 1~209.0 & 544.9 & 341.62& 65.2 & 8.4\\\midrule
    LightGBM  & 6~926 & 1~204.8 & 545.6 & \textbf{336.10} & 64.4 & 6.6\\
    XGBoost & 6~955 & 1~203.9 & 545.7 & 336.18 & 64.5 & 6.0\\
    EGBM  & 6~953 & 1~203.3 & 546.2 & 336.94 & 63.9 & 5.0\\
    GBM  & 7~094 & 1~207.4 & 546.4 & 342.23 & 64.6 & 9.4\\
    CatBoost  & 6~920 & 1~201.1 & 545.5 & 336.26 & 63.1 & 3.2\\\midrule
    XGBoostLSS  & 6~939 & 1~204.5 & 545.4 & 336.97 & 66.1 & 6.4\\
    NGBoost  & \textbf{6~925} & \textbf{1~203.3} & \textbf{544.0} & 336.93 & 58.0 & \textbf{2.6}\\
    cyc-GBM  & 6~935 & 1~198.9 & 544.7 & 338.80 & \textbf{57.3} & 3.6\\
    XGBoostLSSd  & 7~064 & 1~206.7 & 547.3 & 339.91 & 63.1 & 7.0\\
\bottomrule
\end{tabular}
\label{tab:crps-gamma}
\end{table}

\begin{table}[h!]
\caption{Coverage (in \%) of CI at levels 50\%, 75\%, and 95\% on the test BelgianMTPL set.}
\label{tab:coverage-belgianmtpl}
\centering
\begin{tabular}{lrrrrrr}
\toprule
\multirow{2}{*}{Model}& \multicolumn{3}{c}{Lognormal distribution} & \multicolumn{3}{c}{Gamma distribution} \\ \cmidrule{2-4} \cmidrule{5-7}
 & 50\% & 75\% & 95\% & 50\% & 75\% & 95\%\\ \midrule
 GLM & 30.37 & 72.55& 98.01 & 36.18 & 85.21 & 97.15\\
 GAMLSS & 30.71 & 73.07 & 97.94 & 31.27& 78.16& 96.48\\ \midrule
    LightGBM  & 3.97 & 6.59 & 14.27 & 30.22 & 71.42 & 94.86 \\
    XGBoost & 2.21 & 4.04 & 6.82 & 29.33 & 75.36 & 96.25 \\
    EGBM  &  2.28 & 4.12 & 7.12 & 29.63 & 75.28 & 96.37 \\
    GBM  &  1.87 & 3.63 & 6.40 & 27.38 & 74.79 & 96.48 \\
    CatBoost  & 2.13 & 3.97 & 6.67  & 29.74 & 75.24 & 96.22 \\\midrule
    XGBoostLSS  &27.90 & 65.62 & 95.33 & 29.51 & 77.49 & 96.67  \\
    NGBoost  & 28.73 & 73.33 & 97.83 & 29.85 & 78.61 & 96.59 \\
    cyc-GBM  &17.68 & 41.39 & 82.55  & 30.00 & 77.64 & 96.52 \\
    XGBoostLSSd  &  29.51 & 71.95 & 97.45 & 31.42 & 75.62 & 95.96 \\
\bottomrule
\end{tabular}
\end{table}

\begin{table}[h!]
\caption{Coverage (in \%) of CI at levels 50\%, 75\%, and 95\% on the test WorkComp set.}
\centering
\begin{tabular}{lrrrrrr}
\toprule
\multirow{2}{*}{Model}& \multicolumn{3}{c}{Lognormal distribution} & \multicolumn{3}{c}{Gamma distribution} \\ \cmidrule{2-4} \cmidrule{5-7}
 & 50\% & 75\% & 95\% & 50\% & 75\% & 95\%\\ \midrule
 GLM & 53.04 & 77.61 & 94.38 & 47.71& 89.48& 98.54\\
 GAMLSS &45.81 &73.82 & 95.87& 63.33&89.56& 95.58\\ \midrule
    LightGBM  & 57.90 & 80.36 & 94.68 & 56.75 & 81.92 & 93.80 \\
    XGBoost & 55.57 & 78.02 & 93.84 & 60.96 & 87.32 & 95.19 \\
    EGBM  & 55.96 & 78.83 & 94.21 & 61.85 & 87.31 & 95.17 \\
    GBM  & 50.85 & 74.23 & 92.72 & 54.01 & 85.61 & 95.38 \\
    CatBoost  & 40.38 & 60.40 & 84.56 & 60.25 & 86.25 & 94.78 \\\midrule
    XGBoostLSS  & 47.08 & 70.91 & 89.82 & 61.74 & 87.21 & 94.71 \\
    NGBoost  & 52.50 & 77.21 & 93.84 & 59.91 & 87.77 & 94.82 \\
    cyc-GBM  & 45.16 & 69.02 & 88.62 & 58.91 & 88.41 & 95.12 \\
    XGBoostLSSd  &  48.35 & 69.96 & 90.85 & 60.28 & 86.42 & 95.23 \\
\bottomrule
\end{tabular}
\end{table}

The CI coverage on the test set for each model at levels 50\%, 75\%, 95\% are presented in Tables~\ref{tab:coverage-belgianmtpl}--\ref{tab:coverage-emcien}. CI coverage is better for the lognormal models in the WorkComp dataset only, which is synthetic.

\begin{table}[h!]
\caption{Coverage (in \%) of CI at levels 50\%, 75\%, and 95\% on the test freMTPL set.}
\centering
\begin{tabular}{lrrrrrr}
\toprule
\multirow{2}{*}{Model}& \multicolumn{3}{c}{Lognormal distribution} & \multicolumn{3}{c}{Gamma distribution} \\ \cmidrule{2-4} \cmidrule{5-7}
 & 50\% & 75\% & 95\% & 50\% & 75\% & 95\%\\ \midrule
    GLM & 62.84 & 77.53 & 92.12 & 73.14& 92.19& 98.45 \\
    GAMLSS & 63.04 & 77.41 & 92.36 & 65.75& 80.07& 95.22 \\ \midrule
    LightGBM  & 16.84 & 27.27 & 44.20 & 62.93 & 77.71 & 94.48 \\
    XGBoost & 11.89 & 21.66 & 37.32 & 65.54 & 79.58 & 95.01 \\
    EGBM  &  7.53 & 15.01 & 29.10 & 65.75 & 79.95 & 95.10 \\
    GBM  & 10.02 & 19.56 & 34.30 & 66.33 & 79.95 & 95.40 \\
    CatBoost  &  7.55 & 16.21 & 30.76 & 66.35 & 79.88 & 95.29 \\\midrule
    XGBoostLSS  & 63.05 & 77.55 & 92.24 & 66.24 & 80.28 & 95.40 \\
    NGBoost  & 61.48 & 76.56 & 91.48 & 65.01 & 79.10 & 94.73 \\
    cyc-GBM  & 58.38 & 73.05 & 88.73 & 59.35 & 74.36 & 90.02 \\
    XGBoostLSSd  & 62.03 & 77.14 & 91.69 & 65.31 & 79.58 & 94.90 \\
\bottomrule
\end{tabular}
\end{table}

\begin{table}[t]
\caption{Coverage (in \%) of CI at levels 50\%, 75\%, and 95\% on the test pg15training set.}
\centering
\begin{tabular}{lrrrrrr}
\toprule
\multirow{2}{*}{Model}& \multicolumn{3}{c}{Lognormal distribution} & \multicolumn{3}{c}{Gamma distribution} \\ \cmidrule{2-4} \cmidrule{5-7}
 & 50\% & 75\% & 95\% & 50\% & 75\% & 95\%\\ \midrule
 GLM & 53.37 & 79.82 & 95.21 & 51.08& 75.30& 94.78\\
 GAMLSS & 49.40 & 74.21 & 92.87 & 47.77& 71.38& 91.89 \\ \midrule
    LightGBM  & 12.40 & 20.62 & 35.47 & 49.78 & 74.43 & 93.74\\
    XGBoost & 11.31 & 19.59 & 33.46 & 50.05 & 74.37 & 93.42\\
    EGBM  & 11.05 & 19.37 & 34.28 & 49.84 & 74.10 & 93.63\\
    GBM  & 7.02 & 10.55 & 20.00 & 50.82 & 75.19 & 93.91\\
    CatBoost  & 12.57 & 20.84 & 35.15 & 49.84 & 74.92 & 93.63\\\midrule
    XGBoostLSS  & 51.58 & 78.56 & 94.56 & 51.03 & 75.84 & 94.06\\
    NGBoost  & 51.25 & 78.02 & 94.34 & 49.78 & 73.50 & 93.47\\
    cyc-GBM  & 49.78 & 74.70 & 93.15 & 47.61 & 71.82 & 92.22\\
        XGBoostLSSd  & 47.44 & 72.47 & 91.57 & 45.97 & 70.24 & 91.57\\
\bottomrule
\end{tabular}
\end{table}

\begin{table}[t]
\caption{Coverage (in \%) of CI at levels 50\%, 75\%, and 95\% on the test Emcien set.}
\label{tab:coverage-emcien}
\centering
\begin{tabular}{lrrrrrr}
\toprule
\multirow{2}{*}{Model}& \multicolumn{3}{c}{Lognormal distribution} & \multicolumn{3}{c}{Gamma distribution} \\ \cmidrule{2-4} \cmidrule{5-7}
 & 50\% & 75\% & 95\% & 50\% & 75\% & 95\%\\ \midrule
    GLM & 77.60 & 87.93 & 95.73 & 64.87 & 83.33& 92.13\\
    GAMLSS & 50.60 & 78.47 & 94.67 & 58.24&82.37& 95.23\\ \midrule
    LightGBM  & 89.13 & 95.33 & 98.67 & 68.87 & 84.07 & 92.93\\
    XGBoost & 88.53 & 95.27 & 98.73 & 71.00 & 85.73 & 94.20\\
    EGBM  & 88.27 & 95.40 & 98.67 & 69.67 & 84.87 & 93.47\\
    GBM  & 88.27 & 95.53 & 98.60 & 71.00 & 85.47 & 94.67\\
    CatBoost  & 99.80 & 100.00 & 100.00 & 69.80 & 84.93 & 93.33\\\midrule
    XGBoostLSS  & 47.53 & 75.47 & 93.93 & 79.73 & 92.00 & 98.73\\
    NGBoost  & 44.20 & 71.20 & 93.00 & 50.80 & 75.53 & 95.00\\
    cyc-GBM  & 38.00 & 58.13 & 78.67 & 36.33 & 57.67 & 78.20\\
    XGBoostLSSd  & 43.73 & 73.40 & 93.87 & 73.67 & 89.60 & 98.40\\
\bottomrule
\end{tabular}
\label{tab:coverage-lognorm}
\end{table}

We compare the tariff structures through the Gini index in Tables~\ref{tab:giniindex_poissongamma} and~\ref{tab:giniindex_poissongammafrench}.

\begin{landscape}

\vspace*{\fill}
\begin{table}[ht]
    \centering
       \caption{Pairwise comparison of the tariff structures resulting from the Poisson frequency and gamma severity rebalanced models on the test BelgianMTPL set according to the Gini Index. Row-wise maximal values are in bold.}
\begin{tabular}{lrrrrrrrrrrr}
    \toprule
    & GLM & GBM & XGBoost & LightGBM & CatBoost & NGBoost & XGBoostLSS & XGBoostLSSd & cyc-GBM & EGBM & GAMLSS\\
    \midrule
    GLM           & & 11.91 & 12.68 & 12.35  & \textbf{13.68}  & 10.19  & 2.87  & 9.85  & 12.22 & 12.41 & 11.48\\
    GBM           & -3.10 &  & 0.26   & -1.40  & -1.33 & \textbf{2.77}   & -4.27 & -4.54 & 0.27   & -0.97 & 2.65\\
    XGBoost       & -2.32 & \textbf{5.73}   &  & 2.19   & 2.51   & 1.29   & -6.37 & -3.12& 1.77   & 2.80  & 5.60\\
    LightGBM      & -1.47 & \textbf{7.22}   & 2.40   &    & 3.82   & 3.74   & -3.79 & -1.25 & 4.20   & 2.93  & 7.14\\
    CatBoost      & -4.10 & \textbf{6.06}   & 2.01   & 1.48   &    & 3.33   & -4.60 & -2.13 & 1.65   & 2.16  & 5.67\\
    NGBoost       & 6.27  & \textbf{11.37}  & 9.91   & 8.75   & 9.91   &   & 2.91  & 7.07  & 10.16 & 10.29 & 11.10\\
    XGBoostLSS    & 10.06 & 18.18  & 17.74  & 16.84  & 17.13  & 12.89  & & 16.68 & \textbf{18.87}  & 16.49 & 18.22\\
    XGBoostLSSd   & 0.14  & \textbf{12.17}  & 9.02  & 8.48   & 8.63   & 6.68   & -8.92 &  & 7.24   & 7.93  & 11.09\\
    cyc-GBM       & -1.76 & \textbf{7.00}   & 2.53   & 1.89   & 4.25   & 2.07   & -8.98 & -1.58 &  & 2.90  & 6.16\\
    EGBM          & -0.23 & \textbf{7.69}   & 3.84   & 2.90   & 4.20  & 2.89   & -1.76 & 1.70  & 4.85   &  & 7.09\\
     GAMLSS        & -5.80 &  \textbf{5.16} &  3.46  &  2.94 & 3.10  &  2.51  & -3.24 &  0.38 &  4.37  & 3.00 &  \\
    \bottomrule
    \end{tabular}
    \label{tab:giniindex_poissongamma}
\end{table}
\vspace*{\fill}

\vspace*{\fill}
\begin{table}[ht]
    \centering
       \caption{Pairwise comparison of the tariff structures resulting from the Poisson frequency and gamma severity rebalanced models on the test freMTPL set according to the Gini Index. Row-wise maximal values are in bold.}
\begin{tabular}{lrrrrrrrrrrr}
    \toprule
    & GLM & GBM & XGBoost & LightGBM & CatBoost & NGBoost & XGBoostLSS & XGBoostLSSd & cyc-GBM & EGBM & GAMLSS \\
    \midrule
    GLM           &   & 18.47 & \textbf{28.98}  & 19.66  & 20.26  & 13.76  & 13.35  & 10.76  & 22.49 & 18.44 & 11.14\\
    GBM           & 8.56 &    & \textbf{22.93}  & 9.69  & 4.00  & -1.09   & 11.08 & 9.60 & 16.19   & 1.87 & 10.62 \\
    XGBoost       & 27.76 & 26.67   &  & 26.99   & 27.82  & 23.78   & 29.76 & 29.18 & \textbf{33.17}   & 26.42 & 27.71\\
    LightGBM      & 6.42 & -3.52   & \textbf{24.04}   &    & 0.44   & -3.41   & 9.55 & 8.21 & 15.15   & -1.00  & 8.39 \\
    CatBoost      & 4.99 & 5.08   & \textbf{21.76}   & 9.68   &   & -0.07   & 7.68 & 6.07 & 15.99   & 6.45  & 17.75\\
    NGBoost       & 12.89  & 17.54 & \textbf{27.47}   & 19.34   & 16.83   &  & 14.57  & 13.81  & 24.47  & 12.86 & 17.75 \\
    XGBoostLSS    & 3.33 & 16.42  & \textbf{29.27}  & 17.52  & 18.33  & 7.41  &  & -9.90 & 21.27  & 16.58 & 10.42 \\
    XGBoostLSSd   & 5.90  & 17.78  & \textbf{29.86}  & 18.87   & 19.80  & 8.90   & 12.62 &   & 22.16   & 17.90  & 11.99 \\
    cyc-GBM       & 6.31 & 4.06   & \textbf{24.21}  & 5.42  & 4.77   & 0.87   & 5.93 & 5.56 &   & 3.72  & 10.10 \\
    EGBM          & 9.43 & 10.33   &\textbf{ 24.14}  & 13.83   & 6.59  & 6.13  & 12.89 & 11.64 & 18.29   &   & 11.31 \\
    GAMLSS         & 3.47  & 15.56  & \textbf{26.20} & 16.25  & 16.74 & 9.38 & 10.02 & 8.63 &  19.34  & 15.58 &  \\
    \bottomrule
\end{tabular}
    \label{tab:giniindex_poissongammafrench}
\end{table}
\vspace*{\fill}
\end{landscape}

\end{document}